\documentclass[letterpaper,journal]{IEEEtran}

\IEEEoverridecommandlockouts                              

\usepackage{amsthm}
\usepackage{graphicx}
\usepackage{caption}
\usepackage{subcaption}
\usepackage[linesnumbered,ruled,vlined]{algorithm2e}
\usepackage{algorithmicx}
\usepackage{amsmath}
\theoremstyle{plain}
\usepackage{mathtools}
\usepackage{epsfig}
\usepackage{amssymb}
\usepackage{epstopdf}
\usepackage{multicol}
\usepackage{multirow}
\usepackage{array}
\usepackage{url}
\usepackage{color}
\usepackage{float}
\usepackage{wrapfig}
\usepackage{placeins}
\usepackage{booktabs}
\def\BibTeX{{\rm B\kern-.05em{\sc i\kern-.025em b}\kern-.08em
    T\kern-.1667em\lower.7ex\hbox{E}\kern-.125emX}}

\usepackage{blindtext}
\usepackage{nidanfloat}

\title{\LARGE \bf
Constrained Motion Planning Networks X
}

\author{Ahmed H. Qureshi, Jiangeng Dong, Asfiya Baig and Michael C. Yip
\thanks{A. H. Qureshi, J. Dong, A. Baig and M. C. Yip are affiliated with University of California San Diego, La Jolla, CA 92093 USA. {\tt\small \{a1qureshi, jid103, abaig, yip\}@ucsd.edu}}%
}

\begin{document}
%

\maketitle
\thispagestyle{empty}
\pagestyle{empty}


\begin{abstract}
Constrained motion planning is a challenging field of research, aiming for computationally efficient methods that can find a collision-free path on the constraint manifolds between a given start and goal configuration. These planning problems come up surprisingly frequently, such as in robot manipulation for performing daily life assistive tasks. However, few solutions to constrained motion planning are available, and those that exist struggle with high computational time complexity in finding a path solution on the manifolds. To address this challenge, we present Constrained Motion Planning Networks X (CoMPNetX). It is a neural planning approach, comprising a conditional deep neural generator and discriminator with neural gradients-based fast projection operator. We also introduce neural task and scene representations conditioned on which the CoMPNetX generates implicit manifold configurations to turbo-charge any underlying classical planner such as Sampling-based Motion Planning methods for quickly solving complex constrained planning tasks. We show that our method finds path solutions with high success rates and lower computation times than state-of-the-art traditional path-finding tools on various challenging scenarios.
\end{abstract}

\section{INTRODUCTION}
Constrained Motion Planning (CMP) has a broad range of robotics applications for solving practical problems emerging in domains such as assistance at home, factory floors, disaster sites, and hospitals~\cite{choset2005principles}. In our daily life, most of our activities involve a large number of CMP tasks. For example, at our home, we interact with various objects to perform usual household chores such as cleaning and cooking, including opening doors, carrying a  tray or a glass filled with water, and lifting boxes. Likewise, skilled workers manipulate their tools to solve a wide variety of tasks such as assembly at factory floors and advanced-level surgery in the hospitals. 

In all of the scenarios mentioned above, our cognitive process decomposes a given task (e.g., cleaning) into subtasks (e.g., moving objects to their designated places) and accomplishes them sequentially or concurrently by sending motor commands to the body for physical interaction with the environment under the task-specific constraints~\cite{cooper2000contention, zacks2007event}. In robotics, this phenomenon is known as Task and Motion Planning (TMP). A task planner decomposes a given task into a sequence of sub-tasks, and a motion planner achieves those sub-tasks by planning feasible robot motion sequences. This paper focuses on the latter part of TMP, i.e., task-constrained motion planning methods, and their integration with the existing state-of-the-art learning-based task programmers.

\begin{figure}[t]
    \centering
    \begin{subfigure}[b]{0.24\textwidth}
       \includegraphics[width=4.75cm]{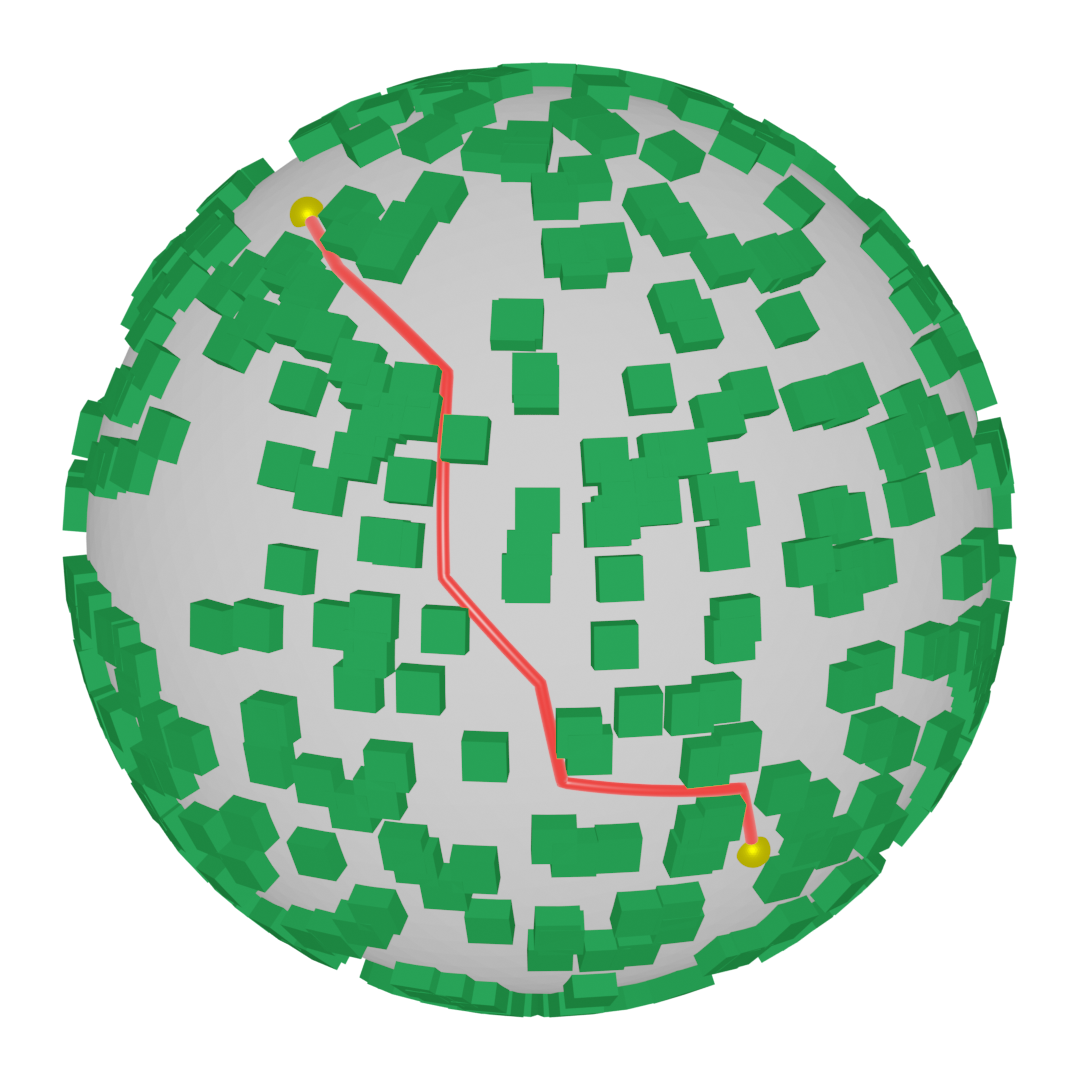}
       \caption{}
    \end{subfigure}    
    \begin{subfigure}[b]{0.24\textwidth}
     \includegraphics[width=4.75cm]{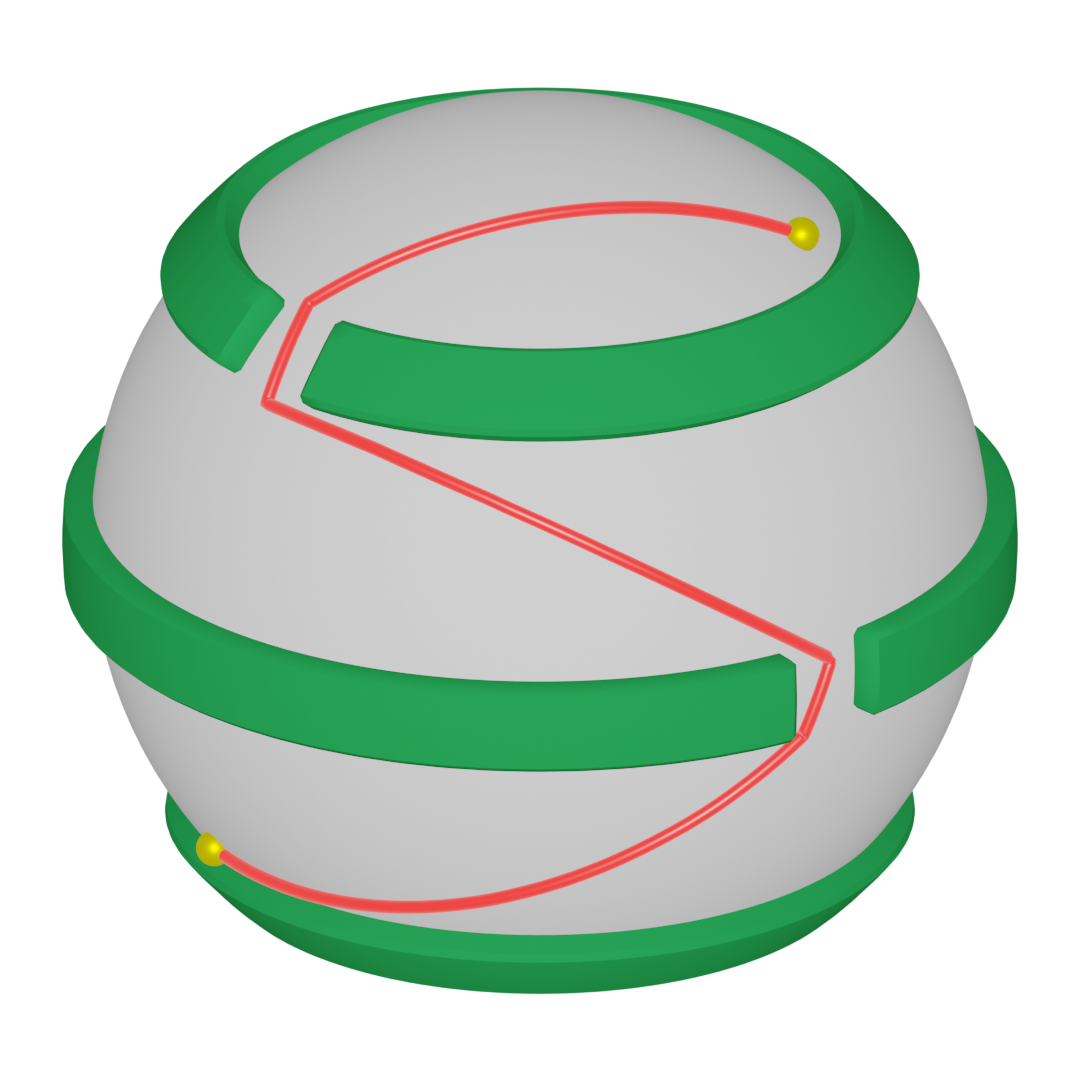}
     \caption{}
    \end{subfigure}
    \caption{CoMPNetX generalized in sphere environment from (a) small cubical obstacles' geometry to (b) multiple longitudinal obstacle strips and planned near-optimal paths between randomly selected start and goal pairs in sub-second computational times.}\label{sphere1}
\end{figure}
In the last decade, Sampling-based Motion Planning (SMP) methods have surfaced as prominent motion planning tools in robotics~\cite{lavalle2006planning}. These algorithms randomly sample the robot joint-configurations to build a collision-free graph, which eventually connects the given start and goal configurations leading to a path solution~\cite{lavalle2006planning}. However, in CMP, the constraint equations implicitly define a configuration space comprising zero-volume constraint manifolds embedded in a higher-dimensional ambient space of the robot's joint variables~\cite{jaillet2013efficient}. Therefore, the probability of generating random robot configurations on those manifolds is not just low but zero, which makes the state-of-the-art gold standard SMP methods~\cite{kuffner2000rrt,karaman2011sampling,gammell2014informed} \cite{gammell2015batch,qureshi2015intelligent,janson2015fast,qureshi2016potential,tahir2018potentially} fail in such problems~\cite{kingston2018sampling}. 

Recently, constraint-adherence methods that generate samples on the manifolds have been incorporated into existing SMP algorithms for CMP~\cite{kingston2018sampling}. These methods include projection and continuation-based approaches. The former uses Jacobian-based gradient descent to project a given configuration to the manifold. The latter takes a known constraint-adhering configuration to compute a tangent space using which new samples are generated closer to the manifold for projection. These advanced planning methods solve a wide range of tasks, but they often exhibit high computational time complexity with high variance, making them frequently impractical for real-world manipulation problems. 

A parallel development led to the cross-fertilization of SMP and machine-learning approaches, resulting into learning-based motion planners~\cite{qureshi2019motion,qureshi2019motionb,ichter2018learning,bency2019neural,qureshi2018deeply,ichter2019robot,johnson2020dynamically}. These methods learn from an oracle planner and are shown to be scalable and generalizable to new problems with significantly faster computational speed than classical methods. Some of these planners even provide worst-case theoretical guarantees. For instance, Motion Planning Networks (MPNet)~\cite{qureshi2019motion,qureshi2019motionb} generates collision-free paths through divide-and-conquer as it divides the problem into sub-problems and either replans or outsources them, in worst-case, to a classical planner while still retaining its computational benefits. 

In our recent work, we extended MPNet to solve CMP problems by proposing Constrained Motion Planning Networks (CoMPNet)~\cite{qureshi2020compnet}. CoMPNet is a deep neural network-based approach that takes the environment perception information, text-based task specification defining the constraints  (e.g., \textit{open the door}), and robot's start and goal configurations as an input and outputs a feasible path on the constraint manifolds. CoMPNet connects any two given configurations using a projection-based constraint-adherence operator, and like MPNet, it also performs a divide-and-conquer through bidirectional expansion. However, it avoids replanning, which is a computationally expensive process in CMP, and instead builds an informed tree of possible paths.
  
This paper presents a unified framework called Constrained Motion Planning Networks X (CoMPNetX)\footnote{The project videos and other supplementary material are available at https://sites.google.com/view/compnetx/home}, which extends CoMPNet and generates informed implicit manifold configurations to speed-up any SMP algorithm equipped with their constraint-adherence approach for solving CMP problems. CoMPNetX comprises the conditional neural generator, discriminator, a neural gradient-based projection operator, and sampling heuristics to propose samples for all kinds of SMP methods. Furthermore, compared to our previously proposed CoMPNet, this new approach, i.e., CoMPNetX, has the following novel features:
\begin{itemize}
    \item CoMPNetX plans in implicit manifold configuration spaces, whereas CoMPNet only considers the robot configuration space. The implicit manifold configuration spaces are formed by the robot configuration and the constraint function. For instance, in the door opening task,  the door, represented as a virtual-link manipulator using Task Space Regions (TSRs), and the robot arm forms an implicit manifold planning space for CoMPNetX.
    \item CoMPNet only considers the projection operator for constraint adherence. In contrast, in this paper, we extend CoMPNet, naming it CoMPNetX, to operate with both projection- and continuation-based constraint adherence approaches for enhancing any SMP method, including batch and bidirectional techniques.
    \item In our previous work, the task sequences were defined by an expert as a text, e.g., open the cabinet and then move an object into the cabinet. CoMPNet sequentially takes the latent embeddings of those text-based task specifications to generate the motion sequences. However, text-based representations are agnostic of the given workspace and the overall planning objective. Therefore, this paper introduces a strategy to combine CoMPNetX with the deep neural network-based task planning approaches that relieve an expert from providing task sequences during execution and provide context-aware neural task representations for CMP.
    \item Unlike CoMPNet, the proposed approach also comprises a discriminator function that predicts the distances of generated configurations from the constraint manifold and provides gradients to project them to the manifold if needed.
\end{itemize}
In summary, CoMPNetX can generate robot configurations for a wide range of SMP algorithms while retaining their worst-case theoretical guarantees. Our generator and discriminator are conditioned on the neural task representation and the environment observation encoding. The conditional generator takes the desired start and goal configurations to output intermediate implicit manifold configurations, and the conditional discriminator predicts their geodesic distances from the underlying manifold. We use the discriminator's predictions and their gradients as the operator to project the given configurations towards the constraint manifold if needed. CoMPNetX naturally forms a mutual symbiotic relationship with learning-based task programmers and exploits their inner states, representing tasks, to transverse multiple constrained manifolds for finding their path solutions. We show that these task representations from a learning-based task planner can lead to better performance in motion planning than human-defined text-based task representations (as in \cite{qureshi2020compnet}). We test CoMPNetX with various SMP algorithms using both continuation and projection-based constraint-adherence methods on challenging problems and benchmark them against the state-of-the-art classical CMP algorithms. We also evaluate our models' generalization capacity to new planning problems and environment structures, such as in the sphere environment from being trained on settings with small obstacle blocks and generalizing to the environment with multiple obstacle strips forming various narrow passages (Fig.~\ref{sphere1}).   

The remainder of the paper is organized as follows. Section II presents preliminaries describing general notations and ideas in CMP, such as constraint functions and their constraint-adherence methods. Section III offers a detailed literature review on existing approaches in CMP. Sections IV describes our procedure to obtain neural task representations, and Section V presents CoMPNetX with its batch and bidirectional sampling heuristics. Section VI gives implementation details followed by Section VII which is dedicated to experimental results of our comparison, ablation, and extended studies. Section VIII presents a brief discussion about our method inheriting an underlying SMP algorithm's worst-case theoretical properties. Finally, Section IX concludes our work with pointers to our future directions, and an Appendix provides details on the model architectures, algorithmic implementations, and their related parameters.

\section{Preliminaries}
In this section, we describe the problem of constrained motion planning with its basic terminologies. We also outline a brief overview of constrained-adherence operators employed by CMP methods for local planning under hard kinematic constraints.
\subsection{Problem Definition}
In the classical problem of motion planning, the robot system is defined by a configuration space (C-space) $\mathcal{Q} \in \mathbb{R}^n$ with $n \in \mathbb{N}$ dimensions. The axis of C-space corresponds to the system's variables that govern their motion, such as robot joint-angles, and hence, the dimension $n$ is equivalent to the robot's degree-of-freedoms (DOF). The robot's surrounding environment is usually described as task-space $\mathcal{X} \in \mathbb{R}^m$ with $m \in \mathbb{N}$ dimensions, comprising obstacle $\mathcal{X}_{obs} \subset \mathcal{X}$ and obstacle-free $\mathcal{X}_{free}=\mathcal{X} \backslash \mathcal{X}_{obs}$ spaces. In the C-space terminology, the spaces $\mathcal{X}_{obs}$ and $\mathcal{X}_{free}$ are represented as $\mathcal{Q}_{obs}$ and $\mathcal{Q}_{free}= \mathcal{Q} \backslash \mathcal{Q}_{obs}$, respectively. In motion planning, a collision-checker $\mathrm{InCollision}(\cdot)$ is assumed to be available that takes a robot configuration $\boldsymbol{q} \in \mathcal{Q}$ and $\mathcal{X}_{obs}$, and outputs a boolean indicating if a given configuration lies in $\mathcal{Q}_{obs}$ or not. 

We consider a setup where for a given current $\boldsymbol{x}_t \in \mathcal{X}_{free}$ and target $\boldsymbol{x}_T \in \mathcal{X}_{free}$ workspace observations, the high-level task planner, $\pi_{H}$, at time $t$, outputs an achievable sub-task representation $\boldsymbol{Z}_c$ for the low-level agent $\pi_L$. For each subtask, $\boldsymbol{Z}_c$, we also assume there exist a constraint function $\mathbf{F}$. The agent, $\pi_L$, finds motion sequences in $\mathcal{Q}_{free}$ to achieve the given subtask, $\boldsymbol{Z}_c$, under constraints $\mathbf{F}$, leading to a next observation $\boldsymbol{x}_{t+1}$. This paper considers deep neural networks-based state-of-the-art task planners as high-level agents, $\pi_{H}$, and proposes a novel low-level agent, $\pi_{L}$, i.e., CoMPNetX, that leverages $\{\boldsymbol{Z}_c,\mathbf{F}\}$ for motion planning under task-specific constraints.

A fundamental unconstrained motion planning problem for a given start configuration $\boldsymbol{q}_{init} \in \mathcal{Q}_{free}$, a goal region $\mathcal{Q}_{goal} \subset \mathcal{Q}_{free}$, environment obstacles $\mathcal{X}_{obs}$, and a collision-checker, is defined as:
\\\\
\textbf{Problem 1 (Unconstrained Motion Planning)}\textit{ Given a planning problem $\{\boldsymbol{q}_{init},\mathcal{Q}_{goal},\mathcal{X}_{obs}\}$, and a collision-checker, find a collision-free path solution $\sigma:[0,1]$, if one exists, such that $\sigma_0=\boldsymbol{q}_{init}$, $\sigma_1\in \mathcal{Q}_{goal}$, and $ \sigma [0,1] \mapsto \mathcal{Q}_{free}$. }\\

In the constrained motion planning, a planner also has to satisfy a set of hard constraints defined by a function $\mathbf{F}(\boldsymbol{q}): \mathcal{Q} \mapsto \mathbb{R}^k$, such that $\mathbf{F}(\boldsymbol{q})=\boldsymbol{0}$. The $k \in \mathbb{N}$ denotes the number of constraints imposed on the robot motion, which induces an $(n-k)$-dimensional space embedded in the robot's unconstrained ambient C-space, comprising one or more manifolds $\mathcal{M}$, i.e, 
\begin{equation*}
\mathcal{M}=\{\boldsymbol{q}\in \mathcal{Q}\:|\:\mathbf{F}(\boldsymbol{q})=\boldsymbol{0}\}
\end{equation*}
In practice, a configuration $\boldsymbol{q}$ is assumed to be on the manifold if $\|\mathbf{F}(\boldsymbol{q})\|_2<\varepsilon$, where $\varepsilon>0$ is a tolerance threshold. Furthermore, the obstacle and obstacle-free spaces on the manifolds are denoted as $\mathcal{M}_{free}= \mathcal{M} \cap \mathcal{Q}_{free}$ and $\mathcal{M}_{obs}=\mathcal{M}\backslash \mathcal{M}_{free}$, respectively. A CMP problem for a given start $\boldsymbol{q}_{init}$ configuration, goal region $\mathcal{Q}_{goal} \subset \mathcal{Q}_{free}$, environment obstacles $\mathcal{X}_{obs}$, function $\mathbf{F}$, and a collision-checker, is defined as:  
\\\\
\textbf{Problem 2 (Constrained Motion Planning)}\textit{ Given a planning problem $\{\boldsymbol{q}_{init},\mathcal{Q}_{goal},\mathcal{X}_{obs}, \mathbf{F}\}$, and a collision-checker, find a collision-free path solution $\sigma:[0,1]$, if one exists, such that $\sigma_0=\boldsymbol{q}_{init}$, $\sigma_1\in \mathcal{Q}_{goal}$, and $ \sigma [0,1] \mapsto \mathcal{M}_{free}$.}\\ 

In our work, we show that CoMPNetX solves both unconstrained (Problem 1) and constrained (Problem 2) planning problems. Furthermore, for the latter problem, we only consider kinematic constraints, i.e., the function $\mathbf{F}$ solely depends on robot configuration $\boldsymbol{q} \in \mathcal{Q}$, not on other robot properties such as dynamics representing velocity or acceleration. Moreover, we define $\mathbf{F}(\boldsymbol{q})$ as distance to the constraint manifold with domain $s$, i.e.,
\begin{equation*}
\mathbf{F}(\boldsymbol{q})= \text{Distance to the constraint manifold}
\end{equation*}
For instance, if the constraint is on the robot's end-effector to maintain a particular position, then $\mathbf{F}(\boldsymbol{q})$ can be defined as the distance of the robot's end-effector to that specific position with domain $s \in [0,1]$, spanning an entire or a fraction of a motion trajectory. Likewise, when the robot is moving, balancing constraints are usually imposed on the whole robot motion trajectory with $s=[0,1]$.    

In the remaining section, we describe the two main types of classical constraint-adherence operators that ensure a given configuration or a motion between two configurations lies on the constraint manifold defined by $\mathbf{F}$.
\begin{algorithm}[t]
\DontPrintSemicolon 
\For{$i \gets 0$ \textbf{to} $N$} {
$\Delta \boldsymbol{x} \gets \mathbf{F}(\boldsymbol{q})$ 

\If{$\|\Delta \boldsymbol{x}\|_2 <\varepsilon$}
   {
   \Return{$\boldsymbol{q}$}\;
   	  
   }
 \Else
 {
    $\boldsymbol{q} \gets \boldsymbol{q}- \mathbf{J}(\boldsymbol{q})^+ \Delta \boldsymbol{x}$\;
 }

   }
\caption{Projection Operator: Proj ($\boldsymbol{q}$)}
\label{algo:proj}
\end{algorithm}

\subsection{Projection-based Constraint-Adherance Operator}
The projection operator ($\mathrm{Proj}$) maps a given configuration $\boldsymbol{q} \in \mathbb{R}^n$ to the manifold $\mathcal{M}$. It can be formulated as a constraint optimization problem~\cite{kingston2019exploring}
\begin{align*}
\min_{\boldsymbol{q}'} \cfrac{1}{2} \|\boldsymbol{q}-\boldsymbol{q}'\|^2 
\text{ subject to } \mathbf{F}(\boldsymbol{q}')=\boldsymbol{0},
\end{align*}
with its dual as: 
\begin{equation*}
L(\boldsymbol{q}',\boldsymbol{\lambda})= \cfrac{1}{2} \|\boldsymbol{q}-\boldsymbol{q}'\|^2 -\boldsymbol{\lambda} \mathbf{F}(\boldsymbol{q}'),
\end{equation*}
where $\boldsymbol{\lambda}$ corresponds to Lagrange multipliers. The above system is solved using gradient descent as summarized in Algorithm \ref{algo:proj}, where $\mathbf{J}^+(q)$ is the pseudoinverse of the Jacobian at configuration $\boldsymbol{q} \in \mathcal{Q}$. Algorithm \ref{algo:projsteer} outlines the local planning procedure using a projection operator~\cite{kingston2019exploring,berenson2011task}. This procedure outputs all the intermediate configurations on the manifold in the given conditions and loop limit $N$, when transversing from a given start configuration ($\boldsymbol{q}_s$) towards the end configuration $(\boldsymbol{q}_e)$ in small incremental steps $\gamma \in \mathbb{R}$. The projection-based steering stops if any of the following happens: (i) The loop limit is reached. (ii) The resulting configuration $\boldsymbol{q}_{i+1}$ is in a collision. (iii) The stepping distance is diverging rather than converging to prevent overshooting the target configuration, i.e., either $d_2>d_1$ or $d>\lambda_1\gamma$. (v) The progress in manifold space  $D$ becomes greater than a scalar $\lambda_2$ times the progress in the ambient space $d_w=\|\boldsymbol{q}_e-\boldsymbol{q}_s\|$. 

\begin{algorithm}[h]
\DontPrintSemicolon 
$i \gets 0$; $D \gets 0$ \;
$d_w \gets \|\boldsymbol{q}_e-\boldsymbol{q}_s\|$; $\boldsymbol{q}_i \gets \boldsymbol{q}_{s}$\;
\While{$i<N$} {
$\boldsymbol{q}_{i+1} \gets \mathrm{Proj} (\boldsymbol{q}_i+\gamma(\boldsymbol{q}_{e}-\boldsymbol{q}_i))$\;
$d\gets \|\boldsymbol{q}_{i+1}-\boldsymbol{q}_{i}\|_2$\;
$D \gets D + d$\;
$d_1\gets \|\boldsymbol{q}_{i}-\boldsymbol{q}_{e}\|_2$; $d_2\gets \|\boldsymbol{q}_{i+1}-\boldsymbol{q}_{e}\|_2$\;
\If{$\mathrm{InCollision} (\boldsymbol{q}_{i+1})$ $\boldsymbol{\mathrm{or}}$ $d_2>d_1$ $\boldsymbol{\mathrm{or}}$ $d>\lambda_1 \gamma$ $\boldsymbol{\mathrm{or}}$ $D>\lambda_2 d_w$}
   {
     $\boldsymbol{\mathrm{break}}$\;	  
   }
 }
 $i \gets i+1$\;
 \Return{$\{\boldsymbol{q}_{j}\}^i_{j=0}$}\;

\caption{Projection Integrator ($\boldsymbol{q}_{s},\boldsymbol{q}_{e}$)}
\label{algo:projsteer}
\end{algorithm}
\begin{algorithm}[h]
\DontPrintSemicolon 
$i \gets 0$; $D \gets 0$\; 
$d_w \gets \|\boldsymbol{q}_e-\boldsymbol{q}_s\|$\;
$\boldsymbol{q}_i \gets \boldsymbol{q}_s$\;
$\mathcal{C}_i \gets \mathrm{GetChart}(\boldsymbol{q}_i,\mathcal{A}_\mathcal{M})$\;
$\boldsymbol{u}_i \gets \psi^{-1}_i(\boldsymbol{q}_i)$\;
$\boldsymbol{u}_e \gets \psi^{-1}_i(\boldsymbol{q}_e)$\;
\While{$\|\boldsymbol{u}_i-\boldsymbol{u}_e\|_2>\gamma$} {
$\boldsymbol{u}_{i+1} \gets \boldsymbol{u}_i+\gamma(\boldsymbol{u}_e-\boldsymbol{u}_i)/\|\boldsymbol{u}_e-\boldsymbol{u}_i\|_2$\;
$\boldsymbol{q}_{i+1} \gets \psi_i(\boldsymbol{u}_{i+1})$\;
$d \gets \|\boldsymbol{q}_{i+1}-\boldsymbol{q}_i\|_2$\;
$D \gets D + d$\; 
$d_1\gets \|\boldsymbol{q}_{i}-\boldsymbol{q}_{e}\|_2$; $d_2\gets \|\boldsymbol{q}_{i+1}-\boldsymbol{q}_{e}\|_2$\;

\If{$\mathrm{InCollision} (\boldsymbol{q}_{i+1})$ $\boldsymbol{\mathrm{or}}$ $d_2>d_1$ $\boldsymbol{\mathrm{or}}$ $d>\lambda_1 \gamma$ $\boldsymbol{\mathrm{or}}$ $d<\epsilon$ $\boldsymbol{\mathrm{or}}$ $D>\lambda_2 d_w$ $\boldsymbol{\mathrm{or}}$ $i>N$}
   {
     $\boldsymbol{\mathrm{break}}$\;	  
   }

 $i \gets i+1$\;

\If{$\boldsymbol{\mathrm{not}}$ $\mathrm{RegionValidity}(\boldsymbol{u}_i, \boldsymbol{q}_i)$ $\boldsymbol{\mathrm{or}}$ $\boldsymbol{u}_i \notin \mathcal{P}_{i-1}$}
   {
     
	$\mathcal{C}_i \gets \mathrm{GetChart}(\boldsymbol{q}_i, \mathcal{A}_\mathcal{M})$\;
$\boldsymbol{u}_i \gets \psi^{-1}_i(\boldsymbol{q}_i)$\;
$\boldsymbol{u}_e \gets \psi^{-1}_i(\boldsymbol{q}_e)$\;
   }

   }
    \Return{$\{\boldsymbol{q}_{j}\}_{j=0}^i$}\;
\caption{Atlas Integrator ($\boldsymbol{q}_{s},\boldsymbol{q}_{e}, \mathcal{A}_\mathcal{M}$)}
\label{algo:atlassteer}
\end{algorithm}
\begin{figure}[t]
    \centering
    \begin{subfigure}[b]{0.23\textwidth}
     \includegraphics[width=4.8cm]{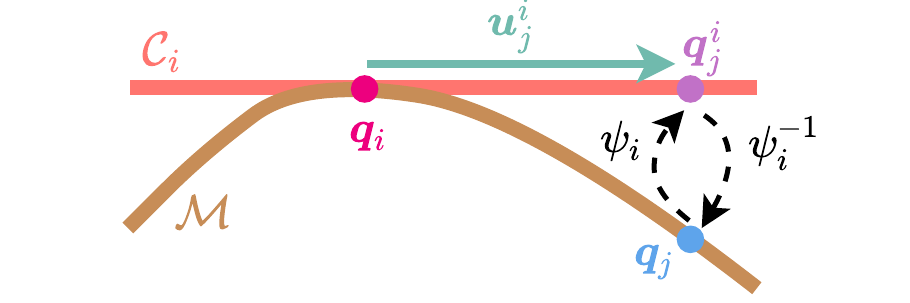}
       \caption{}
    \end{subfigure}
    \begin{subfigure}[b]{0.23\textwidth}
       \includegraphics[width=4.8cm]{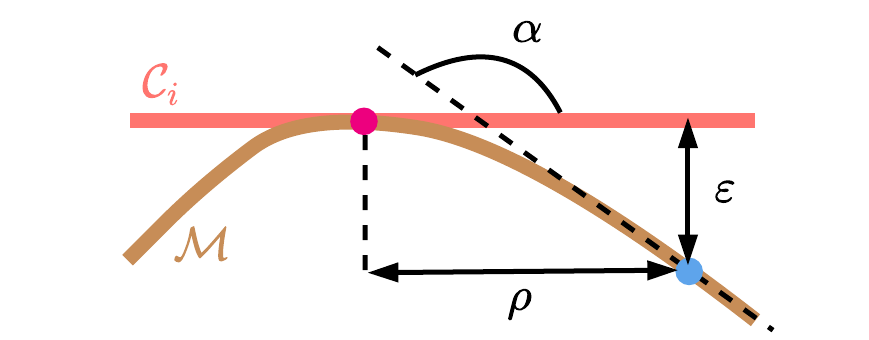}
        \caption{}
    \end{subfigure}
    \caption{(a) A chart $\mathcal{C}_i$ operators comprising exponential $\psi_i$ and logrithmic $\psi^{-1}_i$ functions for mapping between the tangent space at $\boldsymbol{q}_i$ and the manifold. (b) The parameters defining the chart validity region.}\label{tangent}
\end{figure}
\subsection{Continuation-based Constraint-Adherence Operator}
The continuation-based approaches~\cite{kingston2019exploring,jaillet2017path,kim2016tangent} represent the manifold through a set of local parameterizations, known as charts $\mathcal{C}$, forming an \textit{atlas} $\mathcal{A}$.

A chart $\mathcal{C}_i=(\boldsymbol{q}_i,\mathbf{\Phi}_i(\boldsymbol{q}_i))$, with an index $i \in \mathbb{N}$, locally parameterizes a manifold through a tangent space and its orthonormal basis $\mathbf{\Phi}_i$ at a known constraint-adhering configuration $\boldsymbol{q}_i \in \mathcal{M}$. The orthonormal basis $\mathbf{\Phi}_i \in \mathbb{R}^{(n-k)\times n}$ is used to define an exponential map $\psi_i:\mathbb{R}^k \mapsto \mathbb{R}^n$ and its inverse, i.e., a logarithmic map $\psi^{-1}_i:\mathbb{R}^n \mapsto \mathbb{R}^k$, between the parameter $\boldsymbol{u}^i_j$ on the tangent space and the manifold around configuration $\boldsymbol{q}_i$ (Fig. \ref{tangent} (a)).
The basis $\mathbf{\Phi}_i \in \mathbb{R}^{n \times k}$ is computed by solving a following system of equations:     
\begin{equation}
\left( \begin{array}{c} \mathbf{J}(\boldsymbol{q}_i) \\ \mathbf{\Phi}^\top_i \end{array} \right)\mathbf{\Phi}^\top_i = \left( \begin{array}{c} \mathbf{0} \\ \mathbf{I} \end{array} \right),
\end{equation}
where $\mathbf{J}(\boldsymbol{q}_i) \in \mathbb{R}^{k\times n}$ is the Jacobian of $\mathbf{F}$ at the configuration $\boldsymbol{q}_i$, $\mathbf{0} \in \mathbb{R}^{k \times k}$, and $\mathbf{I} \in \mathbb{R}^{k \times k}$ is the identity matrix.

The exponential mapping $\psi_i$ is a two step process. The first step determines a configuration $\boldsymbol{q}^i_j$ in the ambient space using the mapping $\phi_i$, i.e.,
\begin{equation}
\boldsymbol{q}^i_j = \phi_i(\boldsymbol{u}^i_j)=\boldsymbol{q}_i+\mathbf{\Phi}_i \boldsymbol{u}^i_j
\end{equation}
The second step takes the $\boldsymbol{q}^i_j$ and orthogonally projects it to the manifold resulting in $\boldsymbol{q}_j$, by solving the following system:
\begin{equation}
 \left.\begin{aligned}
        \mathbf{F}(\boldsymbol{q}_j)&=\mathbf{0}\\
        \mathbf{\Phi}^\top_i(\boldsymbol{q}_j-\boldsymbol{q}^i_j)&=\mathbf{0}
       \end{aligned}
 \right\}
\end{equation}
The above equations are usually solved iteratively by a Newton method until the error $\|(\boldsymbol{q}_j-\boldsymbol{q}^i_j)\|_2 < \epsilon$ is tolerable or the maximum iteration limit is reached. 

The inverse logarithmic mapping $\psi^{-1}_i$ from the manifold to the tangent space is straightforward to compute, i.e.,   
\begin{equation}
\boldsymbol{u}^i_j=\psi^{-1}_i(\boldsymbol{q}_j)=\mathbf{\Phi}^\top_i(\boldsymbol{q}_j-\boldsymbol{q}_i)
\end{equation}

Note that each chart $\mathcal{C}_i$ has a validity region $\mathcal{V}_i$ in which it properly parameterizes the manifold and exceeding that region could lead to divergence when orthogonaly projecting configurations to the manifold during the exponential mapping process. This validity region is governed by the following conditions:
\begin{equation}
\|\boldsymbol{q}^i_j-\boldsymbol{q}_j\|\leq \varepsilon; \: \: \cfrac{\|\boldsymbol{u}^i_j\|_2}{\|\boldsymbol{q}_i-\boldsymbol{q}_j\|}<\cos(\alpha); \: \: \|\boldsymbol{u}^i_j\|\leq \rho
\end{equation}
where $\varepsilon$ and $\alpha$ indicate the maximum allowable distance and curvature, respectively, between the chart $\mathcal{C}_i$ and the underlying manifold $\mathcal{M}$, and $\rho$ defines the radius of sphere around $\boldsymbol{q}_i$ (Fig. \ref{tangent} (b)). Furthermore, the validity region $\mathcal{V}_i$ can have a complex shape and is usually approximated by a convex polytope $\mathcal{P}_i \subset \mathcal{V}_i$, represented as a set of linear inequalities defined in a tangent space of chart $\mathcal{C}_i$. 

To realize the local planning using continuation operator, there exist two types of methods naming \textit{atlas} integrator (Algorithm \ref{algo:atlassteer}) and \textit{tangent bundle} integrator (Algorithm \ref{algo:tbsteer}). The latter, in contrast to the former, is less strict about the intermediate configurations being on the manifold and performs projections only when needed and does not separate the tangent spaces into half-spaces to prevent overlaps. In our implementations, these integrators assume both start $(\boldsymbol{q}_s)$ and end $(\boldsymbol{q}_e)$ configurations to be on the manifold. The procedure $\mathrm{RegionValididty}$ in the atlas integrator returns $\mathrm{False}$ if any of the above-mentioned region validity conditions are violated. 
  
\begin{algorithm}[h]
\DontPrintSemicolon 
$i \gets 0$; $D \gets 0$\; 
$d_w \gets \|\boldsymbol{q}_e-\boldsymbol{q}_s\|$\;
$\boldsymbol{q}_i \gets \boldsymbol{q}_s$\;
$\mathcal{C}_i \gets \mathrm{GetChart}(\boldsymbol{q}_i,\mathcal{A}_\mathcal{M})$\;
$\boldsymbol{u}_i \gets \psi^{-1}_i(\boldsymbol{q}_i)$\;
$\boldsymbol{u}_e \gets \psi^{-1}_i(\boldsymbol{q}_e)$\;
\While{$\|\boldsymbol{u}_i-\boldsymbol{u}_e\|_2>\gamma$} {
$\boldsymbol{u}_{i+1} \gets \boldsymbol{u}_i+\gamma(\boldsymbol{u}_e-\boldsymbol{u}_i)/\|\boldsymbol{u}_e-\boldsymbol{u}_i\|_2$\;
$\boldsymbol{q}_{i+1} \gets \phi_i(\boldsymbol{u}_{i+1})$\;
$d \gets \|\boldsymbol{q}_{i+1}-\boldsymbol{q}_i\|_2$\;
$D \gets D + d$\;
$d_1\gets \|\boldsymbol{q}_{i}-\boldsymbol{q}_{e}\|_2$; $d_2\gets \|\boldsymbol{q}_{i+1}-\boldsymbol{q}_{e}\|_2$\;

\If{$\mathrm{InCollision} (\boldsymbol{q}_{i+1})$ $\boldsymbol{\mathrm{or}}$ $d_2>d_1$ $\boldsymbol{\mathrm{or}}$ $d>\lambda_1 \gamma$ $\boldsymbol{\mathrm{or}}$ $d<\epsilon$ $\boldsymbol{\mathrm{or}}$ $D>\lambda_2 d_w$ $\boldsymbol{\mathrm{or}}$ $i>N$}
   {
     $\boldsymbol{\mathrm{break}}$\;	  
   }

 $i \gets i+1$\;

\If{$\|\phi_{i-1}(\boldsymbol{u}_i)-\boldsymbol{q}_i\|_2>\varepsilon$ $\boldsymbol{\mathrm{or}}$  $\boldsymbol{u}_i \notin \mathcal{P}_{i-1}$}
   {
     $\boldsymbol{q}_{i} \gets \psi_{i-1}(\boldsymbol{u}_{i})$\;
	$\mathcal{C}_i \gets \mathrm{GetChart}(\boldsymbol{q}_i, \mathcal{A}_\mathcal{M})$\;
$\boldsymbol{u}_i \gets \psi^{-1}_i(\boldsymbol{q}_i)$\;
$\boldsymbol{u}_e \gets \psi^{-1}_i(\boldsymbol{q}_e)$\;
   }

   }
    \Return{$\{\boldsymbol{q}_{j}\}_{j=0}^i$}\;
\caption{Tangent Bundle Integrator ($\boldsymbol{q}_{s},\boldsymbol{q}_{e}, \mathcal{A}_\mathcal{M}$)}
\label{algo:tbsteer}
\end{algorithm}

\section{Related Work}
In this section, we present the existing methods that address the problem of CMP, ranging from relaxation-based methods for trajectory optimization and control to strict approaches such as projection and continuation for sampling-based planning algorithms. 

The relaxation-based methods represent the hard-constraints as soft-constraints by incorporating them as a penalty into the cost function. The cost function is optimized to get the desired robot behavior. For instance, the IK-based reactive control method~\cite{atkeson2015no,johnson2015team} used at the DARPA Robotics Challenge operates in the workspace and finds constrained robot motion through convex optimization of the given cost function. However, these approaches often provide incomplete solutions as they are susceptible to local minima. The trajectory optimization methods~\cite{ratliff2009chomp,schulman2014motion} also optimize the given cost function over the entire trajectory to find a feasible motion plan. However, due to the relaxation, they weakly satisfy the given constraints and are typically only effective on short-horizon problems. Recently, Bonalli et al.~\cite{bonalli2019trajectory} proposed a trajectory optimization method for implicitly-defined constraint manifolds, but their approach is yet to be explored and analyzed in practical CMP robotics problems.

To satisfy hard-constraints without relaxation on the robot motion, the SMP algorithms~\cite{lavalle2006planning}, such as multi-query Probabilistic Road Maps (PRMs)~\cite{kavraki1998probabilistic}, and single-query Rapidly-exploring Random Trees (RRTs)~\cite{lavalle1998rapidly} with its bidirectional variant~\cite{kuffner2000rrt}, have been augmented with constraint-adherence methods, such as projection and continuation, to solve a wide range of CMP problems. 

The projection-based method was first utilized with a variant of PRMs for parallel manipulators under specialized loop-closure constraints~\cite{han00akinematics-based}. The parallel manipulators were treated as active/passive links and were composed into a constraint-adhering configuration using projection. Yakey et. el~\cite{yakey2001randomized} introduced the Randomized Gradient Descent (RGD) method for closed-chain kinematics constraints that generates C-space samples and projects them to the constraint manifold. However, their approach required a significant parameter tuning and was later extended to a generalized framework using RRTs and a Jacobian pseudo-inverse based projection method~\cite{stilman2007task}. In a similar vein, Berenson. et al.~\cite{berenson2011task} proposed the Constrained Bidirectional RRT (CBiRRT) with an intuitive constraint representation approach called Task Space Regions (TSRs). TSRs represent general end-effector pose constraints and allow a quick computation of geodesic distances from the constraint manifolds. Another class of sampling-based methods that use projection operators and plan in the task-space include~\cite{koga1994planning,yamane2004synthesizing,yao2005path}. These methods find a task-space motion plan and find their corresponding configurations in the C-space, which limits their exploration and thus does not yield completeness guarantees. 

The continuation-based methods compute tangent-spaces at a known constraint-adhering configuration to generate new nearby samples for quick projections to the constraint manifold. Yakey et. el~\cite{yakey2001randomized} used continuation to generate new configuration samples within tangent space, which were projected to the manifold using RGD for closed-chain kinematic constraints. The continuation methods have also been used for general end-effector constraints~\cite{weghe2007randomized,stilman2010global}. Inspired by the definition of differentiable manifolds~\cite{spivak1999comprehensive}, recent approaches do not discard tangent spaces. Instead, they compose them using data-structures into an \textit{atlas} for a piece-wise linear approximation of the constraint manifold~\cite{henderson2002multiple}. These methods include Atlas-RRT~\cite{jaillet2017path} and TangentBundle(TB)-RRT~\cite{kim2016tangent} with an underlying single-query bidirectional RRTs algorithm~\cite{kuffner2000rrt}. Atlas-RRT ensures all samples to be on the manifold and separates tangent spaces into tangent polytypes using half-spaces for uniform coverage. In contrast, TB-RRT lazily projects the configurations for constraint-adherence, i.e., only when switching the tangent spaces, and has overlapping tangent spaces, which sometimes lead to invalid states. There also exist variants of Atlas-RRT that allow asymptotic optimality~\cite{jaillet2013asymptotically,jaillet2013efficient} and kinodynamic planning~\cite{bordalba2018randomized} under constraints.

Recently, Kingston et. el~\cite{kingston2019exploring} introduced Implicit MAnifold Configuration Spaces (IMACS) to decouple the choice of constraint-adherence methods from the underlying selection of SMP planners. IMACS highlights that any SMP method equipped with the following two components can solve CMP problems. First, a uniform sampling technique to generate samples on the manifold. Second, a constraint integrator function to connect two configurations on the manifold. IMACS incorporates the constraint function into C-Space, presenting an implicit manifold space to an underlying SMP method. These SMP methods, augmented with a constrained integrator, are shown to solve various CMP problems. Despite these advancements, existing SMP methods are computationally inefficient and take up to several minutes for solving practical problems not just in CMP but also in unconstrained planning problems.

In this paper, we propose CoMPNetX that extends IMACS and our previously proposed CoMPNet~\cite{qureshi2020compnet} and also introduces neural-gradient-based projections to generate informed implicit manifold configurations for underlying SMP methods equipped with any constrained integrator. Our approach can also be interpreted as Neural Informed Implicit MAnifold Configuration Spaces (NIIMACS), which replaces the abstraction layer of IMACS with neural-learned sampling distributions to prioritize sampling in the subsets of a contraint manifold that potentially contains a path solution for a given problem.  

\begin{figure}
    \centering
       \includegraphics[width=8.0cm]{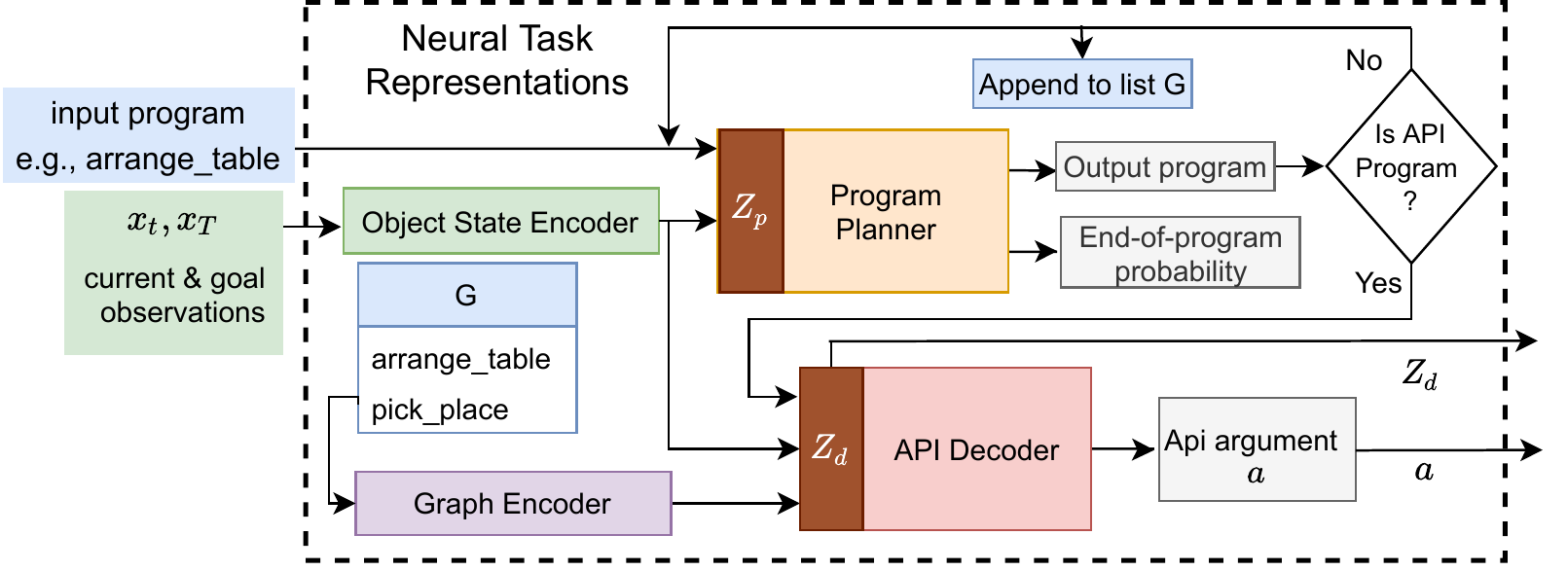}\vspace*{-0.0in}
    \caption{Given a high-level program (e.g., $\mathrm{arrange\_table}$), the environment current $\boldsymbol{x}_t$, and target $\boldsymbol{x}_T$ observations, we obtain the Neural Task Representations for CoMPNetX by exploiting a learning-based task programmer's internal state $\boldsymbol{Z}_d$ and program arguments $\boldsymbol{a}$.}\label{ntp2}
\end{figure}

\section{Neural Task Representations}
This section describes the process to obtain the neural task representations, utilized by CoMPNetX to define task-specific constraints in a scalable and generalizable way. These representations come from the internal state of a learning-based task planner. Although various learning-based task planners can be utilized for acquiring these representations, we adapt a variant \cite{takayanagi2019hierarchical} of the Neural Task Programming (NTP) \cite{xu2018neural} in our framework.

This variant, which we name NTP2, extends original NTP by relieving the need for task demonstration at the test time. NTP2 uses the goal $\boldsymbol{x}_T$ and current $\boldsymbol{x}_t$ observations of the environment to decompose a given high-level task into a feasible sequence of intermediate sub-tasks. We use NTP2 to obtain the neural task representations and the sub-task sequences for CoMPNetX. It comprises the following modules:\\

\textbf{Program Planner:} It is a deep neural network-based iterative program predictor that takes a high-level symbolic task $\boldsymbol{p}_t$, the environment's current $\boldsymbol{x}_t$ and goal $\boldsymbol{x}_T$ observations as an input and outputs a next sub-program $\boldsymbol{p}_{t+1}$ and the end-of-program probability $r$, indicating the accomplishment of a given task.\\

\textbf{API Decoder:} A program is defined as an API program if it requires arguments for the execution. Given an api program $\boldsymbol{p}$ predicted by the program planner, the neural networks based API Decoder predicts their required arguments $\boldsymbol{a}$. The inputs to the API decoder are the current $\boldsymbol{x}_t$ and goal $\boldsymbol{x}_T$ observations, the API program $\boldsymbol{p}$, and a fixed size graph encoding representing the program hierarchy. \\

The overall flow of the algorithm is shown in the Fig. \ref{ntp2}. The current and goal observations are encoded into latent embeddings using their encoders. The program planner, conditioned on observation encodings, iteratively decomposes the given program (e.g., $\mathrm{arrange\_table}$) into subprograms by generating a probability distribution over a set of predefined program instances (e.g., $\mathrm{pick}$ and $\mathrm{place}$). The program with maximum probability is selected, which becomes the input to the program planner in the next iteration. This process is repeated until an API-program is selected. For instance, the given program, $\mathrm{arrange\_table}$, can lead to the selection of a $\mathrm{pick\_place}$ program which subsequently results in the selection of either $\mathrm{pick}$ or $\mathrm{place}$ programs. The $\mathrm{pick}$ and $\mathrm{place}$ are defined as API programs requiring arguments from the API decoder. This API decoder, conditioned on observation encodings and graph embeddings, predicts the API program's arguments indicating the object that needed to be grasped ($\mathrm{pick}$) and moved ($\mathrm{place}$). The graph embeddings are given by the graph encoder that takes a list of non-API programs (Fig. \ref{ntp2}) and encodes them into a fixed-size latent representation. In our implementation of NTP2, the current observation contains the current poses of the given objects in the environment and the robot end-effector pose. The goal observation includes the final poses of all objects at the end of the task. Furthermore, the program planner and the API decoder were trained using the cross-entropy loss for the given expert demonstration. For more details on the implementations, refer to \cite{takayanagi2019hierarchical}, and Appendix A of this paper.

To generate a neural task representation for the CoMPNetX, we take the latent inner embedding $\boldsymbol{Z}_d$ of API Decoder and their corresponding arguments $\boldsymbol{a}$ (Fig. \ref{ntp2}). The internal state $\boldsymbol{Z}_d$ comprises current and goal encodings, graph embedding representing the given task hierarchy, and an API program embedding. Note that the latent state $\boldsymbol{Z}_d$ and arguments $\boldsymbol{a}$ contain sufficient information, i.e., a given high-level task, their sub-task hierarchy, and workspace representation, for the CoMPNetX to effectively plan the feasible robot motion path respecting the task constraints at any instant. This is in contrast to the original CoMPNet framework \cite{qureshi2020compnet} that relied on hand-engineered task plans, and sub-tasks were represented as text-descriptions, making them oblivious of given high-level tasks, their hierarchical structure, and overall workspace setup.
\begin{figure*}[t]
    \centering
       \includegraphics[width=19.0cm]{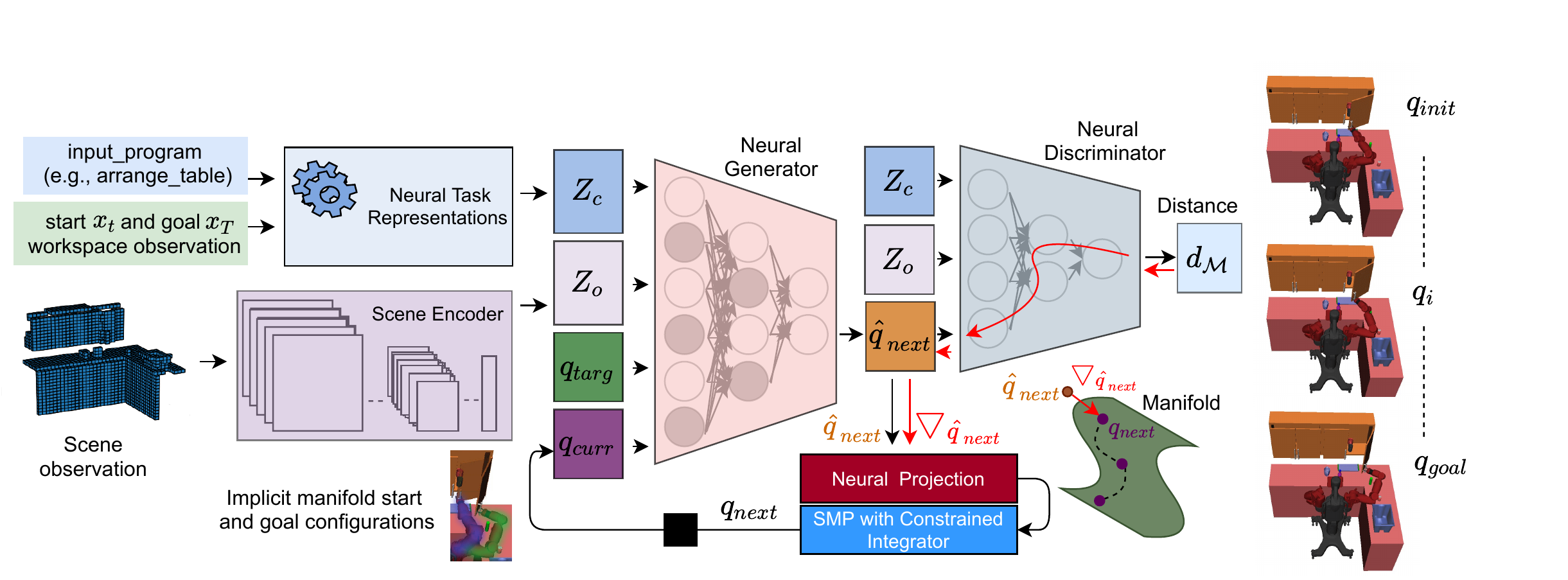}\vspace*{-0.0in}
    \caption{CoMPNetX execution traces for the constrained door opening subtask. Our method comprises a conditional neural generator and discriminator that, in conjunction with a planning algorithm, finds a feasible path solution between start $\boldsymbol{q}_{init}$ (purple) and goal $\boldsymbol{q}_{goal}$ (green) configurations.  
}\label{compnet_exe}
\vspace*{-0.1in}\end{figure*}
\begin{figure*}
\vspace*{-0.0in}
    \centering
       \includegraphics[width=19.0cm]{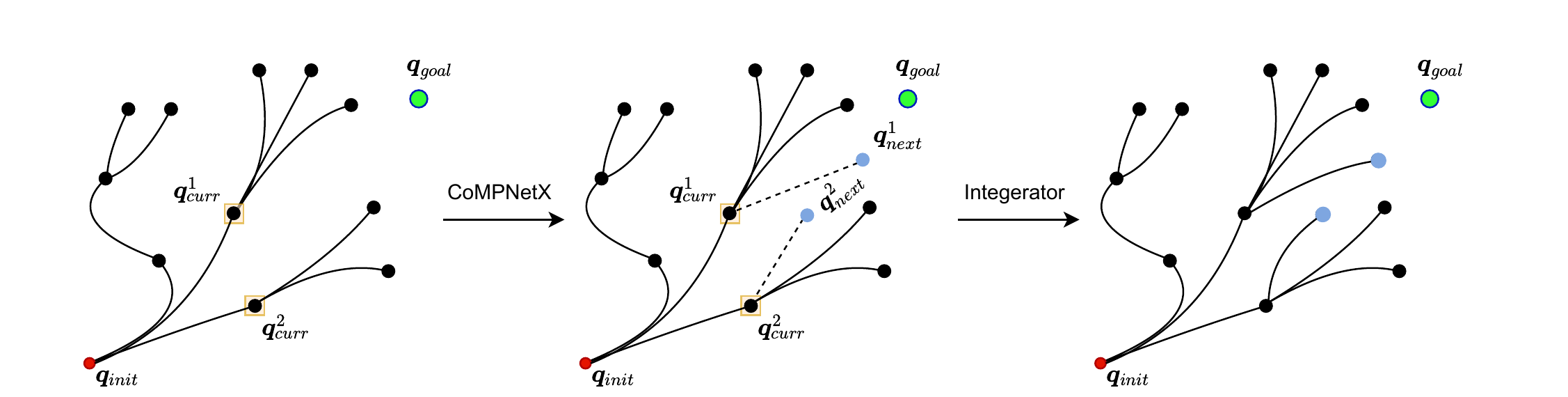}\vspace*{-0.0in}
    \caption{$K$-Batch CoMPNetX: The process shows COMPNetX exploiting neural networks parallelization to generate $K=2$ informed manifold configurations from randomly selected nodes in the tree towards the goal configuration(s) for an underlying SMP method equipped with a constrained-integrator.
}\label{btree}
\vspace*{-0.1in}\end{figure*}

\begin{figure*}[t]
    \centering
    \begin{subfigure}[b]{0.32\textwidth}
       \includegraphics[width=6.8cm]{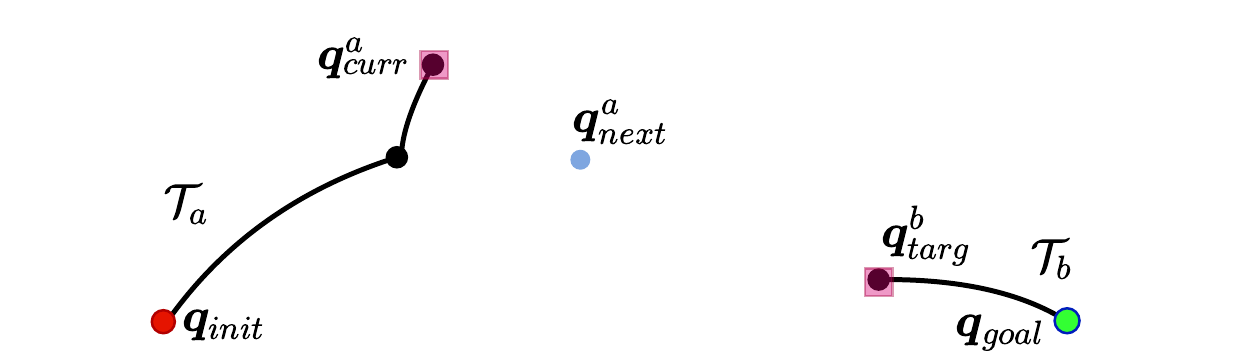}
       \caption{Informed Sample Generation}
    \end{subfigure}    
    \begin{subfigure}[b]{0.32\textwidth}
     \includegraphics[width=6.8cm]{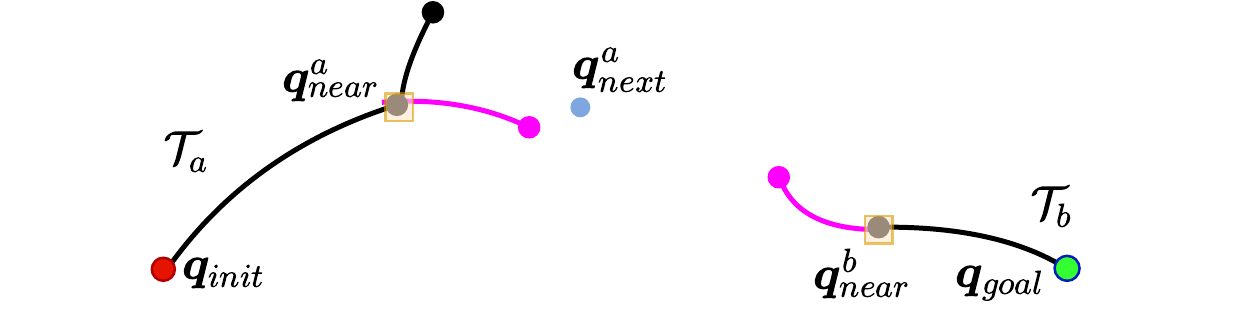}
     \caption{$\mathcal{T}_a$ and $\mathcal{T}_b$ extension}
    \end{subfigure}
    \begin{subfigure}[b]{0.32\textwidth}
       \includegraphics[width=6.8cm]{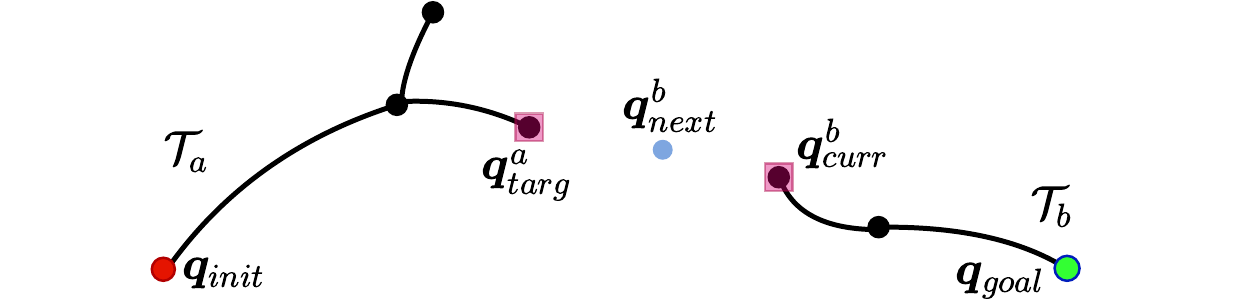}
       \caption{Swapping Roles}
    \end{subfigure}
    \caption{Bidirectional CoMPNetX: (a)-(c) show the CoMPNetX bidirectional sample generation, soliciting neural informed-trees from start and goal to quickly march towards each other within a Bidirectional-SMP method.}\label{bitree}
\end{figure*}
\section{Constrained Motion Planning Networks}
This section formally present CoMPNetX (Fig. \ref{compnet_exe}), comprising a conditional generator, discriminator, neural projection operator, and neural samplers. The neural generator and discriminator are conditioned on the task and scene observation encodings to generalize across different environments and planning problems. Our method with a constrained integrator and an underlying SMP algorithm generates feasible motion plans on the constraint manifolds for the given CMP problems. 

\subsection{Task Encoder}
The task-encoder processes the neural task representations given as $\boldsymbol{Z}_s=[\boldsymbol{Z}_d, \boldsymbol{a}]$. As mentioned earlier, the $\boldsymbol{Z}_d$ is a fixed-sized vector comprising the workspace current and goal observation encodings, the API program embeddings, and the graph encoding (representing the program hierarchy). Our task encoder takes $\boldsymbol{Z}_s$, comprising $\boldsymbol{Z}_d$ and $\boldsymbol{a}$, as an input and composes them into a fixed-size latent embedding $\boldsymbol{Z}_c \in \mathbb{R}^{d_1}$ of size $d_1$ using a neural network.
\subsection{Scene Encoder}
The scene encoder takes the raw environment perception as a 3D depth point-cloud processed into voxels and transforms them to an embedding $\boldsymbol{Z}_o \in \mathbb{R}^{d_2}$ of dimension $d_2$. The 3D voxel grids of dimensions L $\times$ W $\times$ H $\times$ C are converted into 2D voxel patches as L $\times$ W $\times$ (HC), where L, W, H, and C correspond to length, width, height, and the number of channels, respectively. The voxel patches are encoded into $\boldsymbol{Z}_o$ using a 2D convolutional neural network (CNN). We process 3D voxels into 2D voxel patches as 3D maps require 3D-CNNs, which are known to be computationally intensive and their representations often contain empty volumes~\cite{zhang2018efficient}. The scene embedding is passed as a fixed-size feature vector describing the environmental obstacles to a subsequent generator and discriminator. Although neural task representations $\boldsymbol{Z}_c$ contain poses of manipulatable objects in their embeddings, scene observation $\boldsymbol{Z}_o$ also includes information about static non-movable objects acting as obstacles in the environment. 

\subsection{Conditional Neural Generator}
CoMPNetX's generator $G_{\boldsymbol{\phi}}$, with parameters $\boldsymbol{\phi}$, is a stochastic neural model that outputs a variety of implicit manifold configurations leading to a constrained path solution (Fig. \ref{compnet_exe}). Because the generator is trained on both unconstrained and constrained path demonstration data, the output distribution of the neural model tend to fall on or near the constraint manifolds when conditioned on task-specific constraints. Our generator derives its stochastic behavior from using Dropout~\cite{srivastava2014dropout} during inference, which instantly slices $G_{\boldsymbol{\phi}}$ in a probabilistic manner, inculcating variations in the generated samples. Although other techniques such as input Gaussian noise can be used to foster stochasticity, they require a reparametrization trick and are often hard to train end-to-end~\cite{creswell2018generative}. In contrast, Dropout helps capture stochastic behavior from demonstration data, which we observed to be consistently better than hand-crafted input noise distributions in our planning problems. 

The generator's input is the task-observation encodings ($\boldsymbol{Z}_c$ and $\boldsymbol{Z}_o$) that encode the given neural task representations and scene observation, respectively, and the current $\boldsymbol{q}_{curr}$ and target $\boldsymbol{q}_{targ}$ manifold configurations. The output is the next configuration $\hat{\boldsymbol{q}}_{next}$ on/near the constraint manifold that will take the system closer to the given target, i.e.,
\begin{equation}
\hat{\boldsymbol{q}}_{next} \gets G_{\boldsymbol{\phi}}(\boldsymbol{Z}_c,\boldsymbol{Z}_o,\boldsymbol{q}_{curr},\boldsymbol{q}_{targ})
\end{equation}
Given the demonstration trajectories $\sigma^*=\{\boldsymbol{q}^*_0,\cdots,\boldsymbol{q}^*_T\}$ from an oracle planner, we train the generator together with the task and observation encoders in an end-to-end manner using the mean-square loss function, i.e., 
\begin{equation}\vspace*{-0.1in}\label{mse}
\cfrac{1}{N_B} \sum^{N}_{i=0} \sum^{T_i-1}_{j=0} ||\boldsymbol{q}_{i,j+1}-\boldsymbol{q}^*_{i,j+1}||^2,
\end{equation}
where $i$ and $j$ iterates over the number of given paths and the number of nodes in each path, respectively, and $N_B$ is the averaging term. 
\subsection{Conditional Neural Discriminator}
CoMPNetX's discriminator $D_{\boldsymbol{\theta}}$, with parameters $\boldsymbol{\theta}$, is a deterministic neural model that predicts the distance $d_\mathcal{M} \in \mathbb{R}$ of a given configuration $\boldsymbol{\hat{q}}$ from an implicit constraint manifold $\mathcal{M}$ conditioned on the task $\boldsymbol{Z}_c$ and observation $\boldsymbol{Z}_o$ encodings, i.e.,
\begin{equation}
 d_{\mathcal{M}}\gets D_{\boldsymbol{\theta}}(\boldsymbol{\hat{q}},\boldsymbol{Z}_c,\boldsymbol{Z}_o)
\end{equation}

CoMPNetX uses the discriminator predictions and their gradients as the operator, named $\mathrm{NProj}$, to project the given configurations to the constraint manifold if their predicted distances are greater than a threshold $\nu$, thus discriminating samples based on their distances from the manifold  and fixing them accordingly as,
\begin{equation}
\boldsymbol{q}\gets \boldsymbol{\hat{q}} -\gamma \nabla_{\boldsymbol{\hat{q}}}D_{\boldsymbol{\theta}}(\boldsymbol{\hat{q}},\boldsymbol{Z}_c,\boldsymbol{Z}_o),
\end{equation}
where $\gamma \in \mathbb{R}^+$ is a hyperparameter denoting a step size.

To train the discriminator network $D_{\boldsymbol{\theta}}$, we minimize the mean-square loss between its predictions and the true labels. The true labels are the geodesic distances of demonstration trajectories from the constraint manifolds. Furthermore, we introduce a trick to create negative training samples with relatively larger distances from the manifold. The negative training samples comprise the robot configuration from the unconstrained tasks (e.g., reach a given object) and the virtual-link configuration from positive training samples and their corresponding distances are computed by querying $\mathbf{F}$.
\begin{algorithm}[h]
\DontPrintSemicolon 
$\boldsymbol{Z}_c \gets \mathrm{GetTaskEncoding}(\boldsymbol{Z}_s)$\;
$\boldsymbol{Z}_o \gets \mathrm{GetObsEncoding}(\boldsymbol{v})$\;
$\boldsymbol{\hat{q}}_{next} \gets G_{\boldsymbol{\phi}}(\boldsymbol{Z}_c,\boldsymbol{Z}_o,\boldsymbol{q}_{curr},\boldsymbol{q}_{targ})$\;
$d_{\mathcal{M}}\gets D_{\boldsymbol{\theta}}(\boldsymbol{\hat{q}}_{next},\boldsymbol{Z}_c,\boldsymbol{Z}_o)$\;

\If{$d_{\mathcal{M}} > \nu$}
 {
$\boldsymbol{\hat{q}}_{next}\gets \boldsymbol{\hat{q}}_{next} -\gamma \nabla_{\boldsymbol{\hat{q}}_{next}}D_{\boldsymbol{\theta}}(\boldsymbol{\hat{q}}_{next},\boldsymbol{Z}_c,\boldsymbol{Z}_o)$\;
   	  
 }

\Return{$\boldsymbol{\hat{q}}_{next}$}\;
\caption{COMPNetX ($\boldsymbol{Z}_s,\boldsymbol{v},\boldsymbol{q}_{curr},\boldsymbol{q}_{targ}$)}
\label{algo:compnetx}
\end{algorithm}
\subsection{Neural Samplers}
Once trained, CoMPNetX can be used in a number of ways to generate informed neural samples for the underlying SMP algorithms equipped with a constrained adherence method. Fig. \ref{compnet_exe} and Algorithm. \ref{algo:compnetx} present an overall flow of information between different neural modules of CoMPNetX. For a given current $\boldsymbol{q}_{curr}$ and target $\boldsymbol{q}_{targ}$ configuration(s), COMPNetX, conditioned on encodings $\boldsymbol{Z}_{c}$ and $\boldsymbol{Z}_{o}$, generates the next configuration(s) $\boldsymbol{\hat{q}}_{next}$ and projects them towards the constraint manifold using neural gradients if needed. Thanks to CoMPNetX's informed but stochastic sampling and built-in parallelization capacity of neural networks, our method can be adapted to most of underlying SMP methods. For case studies, we present two sampling strategies named $K$-Batch CoMPNetX and Bidirectional CoMPNetX, which together cover a wide range of SMP methods.\\

\begin{algorithm}[h]
\DontPrintSemicolon 
$\mathcal{T} \gets \mathrm{InitializeSMP}(\boldsymbol{q}_{init}, \boldsymbol{q}_{goal})$\;
$K_{\boldsymbol{q}_{targ}} \gets K\mathrm{Replicas}( \boldsymbol{q}_{goal})$\; 
\For{$i \gets 0$ \textbf{to} $N_{max}$} {
\If{$i < N_{ismp}$}
{
$K_{\boldsymbol{q}_{curr}} \gets$ $\mathrm{SelectNodes}(\mathcal{T}, K)$\;
$K_{\boldsymbol{q}_{next}} \gets \mathrm{CoMPNetX}(K_{\boldsymbol{Z}_{s}},K_{\boldsymbol{v}},K_{\boldsymbol{q}_{curr}},K_{\boldsymbol{q}_{targ}})$\;
}
\Else
{
$K_{\boldsymbol{q}_{next}}\gets \mathrm{TraditionalSMP}()$
}
$\mathrm{goal\_reached} \gets \mathrm{SMP}(K_{\boldsymbol{q}_{next}}, \mathcal{T})$\;
\If{$\mathrm{goal\_reached}$}
   {
   $\sigma \gets \mathrm{ExtractPath}(\mathcal{T})$\;
   	  
   }
}
\If{$\sigma$ $\mathrm{is}$ $\mathrm{not}$ $\mathrm{empty}$}
   {
    ExecutePlan($\sigma$)\;
   	  
   }
\Else{
\Return{$\mathrm{Failure}$} or AskExpert\;
}

\Return{$\varnothing$}\;
\caption{$K$-Batch COMPNetX}
\label{algo:btree}
\end{algorithm}
\textbf{$\boldsymbol{K}$-Batch CoMPNetX:}
Our approach exploits the neural networks' innate parallelization capacity to generate a batch of samples with size $K \in \mathbb{N}_{\geq1}$ using CoMPNetX for the underlying unidirectional $(K=1)$ and batch $(K>1)$ SMP methods. In this setup, the input to CoMPNetX is in the form of batches of size $K$. The $K$ target configurations $\boldsymbol{q}_{targ}$ are a set of samples from goal region $\mathcal{G}_{goal}$. The voxel map $\boldsymbol{v}$ and neural task representation $\boldsymbol{Z}_s$ are simply replicated $K$ times. The $K$ current configurations $\boldsymbol{q}_{curr}$ are obtained by randomly selecting $K$ nodes in the graph leading to their corresponding next output configurations as follows: \begin{equation*}
K_{\boldsymbol{q}_{next}}=\mathrm{CoMPNetX}\left(K_{\boldsymbol{Z}_{s}}, K_{\boldsymbol{v}}, K_{\boldsymbol{q}_{curr}}, K_{\boldsymbol{q}_{targ}}\right),
\end{equation*}
\begin{equation}
\textbf{where }
K_{\boldsymbol{q}_{next}}=
\begin{bmatrix}
\boldsymbol{q}^1_{next} \\
\vdots\\
\boldsymbol{q}^K_{next} 
\end{bmatrix},\cdots,
K_{\boldsymbol{q}_{targ}}=
\begin{bmatrix}
\boldsymbol{q}^1_{targ} \\
\vdots\\
\boldsymbol{q}^K_{targ} 
\end{bmatrix}
\end{equation}
At the beginning of planning, the graph $\mathcal{T}$ might have only one sample, i.e., $\boldsymbol{q}_{init}$. In that case, an initial set of $K_{\boldsymbol{q}_{curr}}$ can be created by randomly sampling the manifold $\mathcal{M}_{free}$ or replicating $\boldsymbol{q}_{init}$ for $K$ times. Fig. \ref{btree} shows a case with $K=2$, and Algorithm. \ref{algo:btree} presents $K$-Batch CoMPNetX algorithm with an underlying SMP. This approach is not just for batch sampling methods such as FMT*~\cite{janson2015fast} and BIT*~\cite{gammell2015batch} but can also be applied to any unidirectional SMP method like RRT~\cite{lavalle1998rapidly,kuffner2000rrt} and PRMs~\cite{kavraki1998probabilistic} by setting $K=1$. Furthermore, our procedure shifts to traditional sampling techniques, introduced in IMACS \cite{kingston2019exploring}, after generating neural informed implicit manifold configurations using CoMPNetX for $N_{smp}$ iterations. This allows our framework to explore the entire space in worst-case, leading to theoretical guarantees expected from a planning algorithm. \linebreak

\begin{algorithm}[h]
\DontPrintSemicolon 
$t\gets 1$; $\boldsymbol{p}_0\gets \mathrm{input\_program}$\;
\While{$\boldsymbol{\mathrm{not}}$ $\mathrm{end\_of\_program}$}
{
$\boldsymbol{x}_t,\boldsymbol{v}_t \gets \mathrm{GetObservation()}$\;
$\boldsymbol{p}_t, \boldsymbol{Z}_s, \mathrm{end\_of\_program}  \gets \mathrm{NTP2}(\boldsymbol{x}_t,\boldsymbol{x}_T,\boldsymbol{p}_{t-1})$\;
$\boldsymbol{q}_{init}, \boldsymbol{q}_{goal} \gets \mathrm{GetConfigs(\boldsymbol{p}_t, \boldsymbol{Z}_s)}$\;
$\mathcal{T}_a,\mathcal{T}_b\gets \mathrm{InitializeBiSMP}(\boldsymbol{q}_{init},\boldsymbol{q}_{goal})$\;
$\boldsymbol{q}^a_{curr},\boldsymbol{q}^b_{targ} \gets \boldsymbol{q}_{init}, \boldsymbol{q}_{goal}$\;
\For{$i \gets 0$ \textbf{to} $N_{max}$} {
\If{$i<N_{ismp}$}{
$\boldsymbol{q}^a_{next} \gets \mathrm{CoMPNetX}(\boldsymbol{Z}_s,\boldsymbol{v}_t,\boldsymbol{q}^a_{curr},\boldsymbol{q}^b_{targ})$\;
}
\Else
{
$\boldsymbol{q}^a_{next} \gets \mathrm{TraditonalSMP}()$\;

}
$\boldsymbol{q}^a_{next}, \mathrm{path\_found}   \gets \mathrm{BiSMP}(\boldsymbol{q}^a_{next},\mathcal{T}_a , \mathcal{T}_b)$\;
\If{$\mathrm{path\_found}$}
   {
   $\sigma_t \gets \mathrm{ExtractPath}(\mathcal{T}_a,\mathcal{T}_b)$\;
   	  
   }
   $\boldsymbol{q}^a_{curr} \gets \boldsymbol{q}^a_{next}$\;
	$\mathrm{Swap}(\mathcal{T}_a,\mathcal{T}_b)$\;
	$\mathrm{Swap}(\boldsymbol{q}_{curr},\boldsymbol{q}_{targ})$\;
}
\If{$\sigma_t$ $\mathrm{is}$ $\mathrm{not}$ $\mathrm{empty}$}
   {
    ExecutePlan($\sigma_t$)\;
   	  
   }
\Else{
\Return{$\mathrm{Failure}$} or $\mathrm{AskExpert}$\;
}
$t \gets t+1$\;
}

\Return{$\varnothing$}\;
\caption{Bidirectional COMPNetX}
\label{algo:bitree}
\end{algorithm}
\textbf{Bidirectional CoMPNetX:}
This approach incorporates Bidirectional SMP (BiSMP) methods into CoMPNetX that generate bidirectional trees $\mathcal{T}_a=(V,E)$ and $\mathcal{T}_b=(V,E)$ originating from the start $\boldsymbol{q}_{init}$ and goal $\boldsymbol{q}_{goal}$  configurations, respectively, with vertices $V$ and edges $E$. Although the following approach can be formulated as $K$-Batch bidirectional CoMPNetX, we consider $K=1$ and drop down the $K$ notations introduced in the previous section for brevity. Furthermore, we also show that our approach can be combined with learning-based task planners such as NTP2 that generate neural task representations and intermediate subtasks for CoMPNetX, which in return accomplishes those subtasks, forming a mutually symbiotic relationship. 

In this procedure, CoMPNetX alternatively generates samples for both trees and greedily expands them towards each other by having current and target configurations in the opposite trees (Fig. \ref{bitree}), i.e., 
\begin{equation*}
\textbf{Forward: } \boldsymbol{q}^a_{next} \gets \mathrm{CoMPNetX}\left(\boldsymbol{Z}_{s},\boldsymbol{v}, \boldsymbol{q}^a_{curr}, \boldsymbol{q}^b_{targ}\right)
\end{equation*}
\begin{equation*}
\textbf{Backward: } \boldsymbol{q}^b_{next} \gets \mathrm{CoMPNetX}\left(\boldsymbol{Z}_{s}, \boldsymbol{v}, \boldsymbol{q}^b_{curr}, \boldsymbol{q}^a_{targ}\right)
\end{equation*}
where configurations with superscript $a$ and $b$ corresponds to the tree $\mathcal{T}_a$ and $\mathcal{T}_b$, respectively. \\

Algorithm \ref{algo:bitree} presents an overall framework using NTP2 and CoMPNetX with an underlying bidirectional SMP algorithm, like RRTConnect \cite{kuffner2000rrt}, and a constrained-adherence method. NTP2 takes the current environment observation $\boldsymbol{x}_t$, previous task program $\boldsymbol{p}_{t-1}$, and the desired goal observation $\boldsymbol{x}_T$ and generates the next program $\boldsymbol{p}_t$ with their representation $\boldsymbol{Z}_s$. The procedure $\mathrm{GetConfigs}$ takes the generated task information $(\boldsymbol{p}_t,\boldsymbol{Z}_s)$ and obtains their corresponding start and goal configurations. These configurations and task-scene representations are given to CoMPNetX-BiSMP to accomplish the given subtask by generating a feasible motion plan. 

Fig. \ref{bitree} illustrates the internal process of a BiSMP, such as RRTConnect, using CoMPNetX generated samples. Let's assume tree $\mathcal{T}_a$ current configuration being used to generate the next sample (Fig. \ref{bitree} (a)). The underlying BiSMP begins by extending $\mathcal{T}_a$ towards the next configuration $\boldsymbol{q}^a_{next}$ and updates $\boldsymbol{q}^a_{next}$ with the last state reached by constrained integrator towards the target $\boldsymbol{q}^b_{targ}$ (Fig. \ref{bitree} (b)). The process then extends $\mathcal{T}_b$ towards the $\boldsymbol{q}^a_{next}$ and the extension process ends by returning updated $\boldsymbol{q}^a_{next}$ and a boolean $\mathrm{path\_found}$. The $\mathrm{path\_found}$ is true when trees $\mathcal{T}_a$ and $\mathcal{T}_b$ are connected, depending on trees' connection strategy of an underlying BiSMP, and there exists a path between start and goal that satisfies all the desired constraints. To solicit bidirectional path generation using CoMPNetX, the roles of current and target configurations are also swapped along with planning trees' roles at the end of each planning iteration (Fig. \ref{bitree} (c)). Our CoMPNetX-BiSMP quickly finds a path solution by exploiting the moving targets from its own distribution which improves the stability of the generator to find connectable paths as satisfying the two-point boundary value problem becomes easier when the two goal states are iteratively sampled from a distribution encoded by the generator, rather than one defined arbitrarily during the problem definition.

Note that the constraint function $\mathbf{F}$ is used only by an underlying SMP method. Furthermore like CoMPNet, CoMPNetX (batch and bidirectional) also extends the planning graph from the nearest node of the newly generated next node since all underlying SMP algorithms rely on the nearest neighbor for their graph extension towards the given configuration sample~\cite{lavalle2006planning}. It is also in contrast to the MPNet algorithm~\cite{qureshi2019motion,qureshi2019motionb} that greedily finds a path by extending from $\boldsymbol{q}_{curr}$ to $\boldsymbol{q}_{next}$ in an overall planning method and repairs any non-connectable nodes via stochastic re-planning. Although the MPNet approach works extremely fast in unconstrained planning problems, re-planning becomes computationally expensive in CMP due to projections performed by the constrained integrator. Moreover, the constraint manifolds are non-euclidean in topography, and extension from nearest neighbors becomes convenient for geodesic interpolation. This is evident from the experimentation in our previous work~\cite{qureshi2020compnet}, showing that leveraging MPNet's greedy path-finding approach, without replanning, often fails in finding a connectable path solution on the manifolds. However, in our extended analysis presented in this work, we show that CoMPNetX, in addition to CMP, can still be used with the MPNet planning algorithm for efficiently solving unconstrained planning problems with low computation times and high success rates in high-dimensional planning problems. 

\section{Implementation details}
This section describes the data generation pipeline from setting up scenarios to obtaining expert demonstrations and observation data. We also describe training, and testing data splits for all scenarios considered in this work. Furthermore, with this paper, all generated datasets, trained models, and algorithmic implementations will be made publicly available on our project website\footnote{https://sites.google.com/view/compnetx/home}.

\subsection{Scene setup}
 We setup the following cluttered environments imposing various hard kinematic constraints on the robot motion:
 
\textit{Sphere Environment:} This environment requires the motion planning of a point-mass on the sphere with constraint $\mathbf{F}(\boldsymbol{q})=\|\boldsymbol{q}\|-1$, forming a two-dimensional manifold on a three-dimensional ambient space. In this setup, we create two scenarios:
\begin{itemize}
    \item Scenario 1 - We generate $50$ unique scenes by randomly placing $500$ small obstacle blocks over the sphere (Fig. \ref{sphere1} (a)). For each scene, we randomly sample $2000$ start and goal pairs on the obstacle-free space of the sphere. 
    \item Scenario 2 - This setup requires transversing multiple narrow passages between the randomly selected start and goal configurations (Fig. \ref{sphere1} (b)). We randomly sample the unique $500$ start and goal pairs from the obstacle-free space, each of which constitutes a CMP problem. This setup is only used to test our model's generalization capacity, trained on sphere - scenario 1, to an entirely different environment. 
\end{itemize}

\textit{Bartender Environment:} A dataset, named Bartender environment, containing three different scenarios was created to fully capture the complexities of the real-world task and constrained motion planning problems. The environment includes two tables placed perpendicular to each other. The table contains seven objects placed at random, and only five are movable under pre-specified motion constraints. The five movable/manipulatable items include a juice can (green), fuze bottle (purple), soda can (red), kettle, and red mug. The two stationary objects include a tray and a recycling bin that form the movable objects' goal locations. The juice can, soda can, and fuze bottle are to be placed into the recycling bin with only collision-avoidance constraints. The kettle and the red mug are to be placed on the tray with both stability and collision-avoidance constraints, i.e., no tilting is allowed during the robot motion. The three different scenarios are described as follow.

\begin{itemize}
    \item Scenario 1 - In this scenario, the objects can be moved to their targets in any order. In other words, in most cases, all objects start, and goal configurations are reachable. We generate $1833$ unique scenes through the random placement of the movable and non-movable objects on the tables at the robot's right arm's reachable locations. Each scene contains a total of ten (five unconstrained and five constrained) planning problems.
    
    \item Scenario 2 - In this scenario, the goal location of either the red mug or the kettle contains an obstacle. The obstacles are formed by placing either juice bottle, fuze bottle, or soda can, at the goal location of the kettle or the red mug. For example, if the red mug's goal location contains the juice bottle, the task planner needs to account for this information during the planning process. That is, the juice bottle needs to be moved into the recycling bin before the red mug is attempted to be moved onto the tray. This enforces a constraint on the task planner to account for obstacles. We created 700 scenes in this setup, each with at least two constrained and two unconstrained planning problems.
    
    \item Scenario 3 - In this setup, the kettle and red mug are placed on the tray, and the task is to swap their start locations. In other words, the goal locations of both the kettle and the red mug are occupied by the red mug and the kettle, respectively. Therefore, there is a need for a sub-goal generation for one of the objects. For example, the tea kettle should be moved to a temporary location on the table. This is followed by the pick-place of the plastic mug to its goal location. Finally, the goal location of the tea kettle is now free for its pick-place operation. For this problem, we created 300 unique cases by random placement of the tray, and each case contained atleast six planning problems, i.e., three constrained and three unconstrained. 
\end{itemize} 
\textit{Kitchen Environment:} 
In this scenario we have seven manipulatable objects: soda can, juice can, fuze bottle, cabinet door, black mug, red mug, and pitcher. The objective is to move the cans and bottle to the trash bin, open the cabinet door from any starting angle to a fixed final angle ($\pi/2.7$), transfer (without tilting) the black and red mugs from the cabinet to the tray, and move the pitcher from the table into the cabinet. We construct $1633$ unique scenarios by the random placement of the trash bin, tray, and manipulatable objects (excluding door) on the table and by randomly selecting the cabinet's door starting angle between $0$ to $\pi/4$. Each scenario contains a total of 14 planning problems, i.e., seven unconstrained (reach) and seven constrained (manipulation) problems.
\subsection{Training \& testing data splits}
In the sphere (scenario 1), we use $40$ environments for training and $10$ for testing. The sphere (scenario 2) is used for testing only. In the bartender (scenario 1), and kitchen environments, we use $10\%$ data for testing, and the remainder is used for training. All training paths were generated by an oracle planner, i.e., Atlas-RRTConnect. To train neural task programmer on all bartender (scenarios 1, 2 \& 3) and kitchen environments, we use the same data split ratio, i.e., about  $5\%$ is kept for testing. Note that the CoMPNetX is never trained on the sphere (scenario 2) and Bartender scenarios 2 and 3. We use them to evaluate CoMPNetX generalization capacity across different environment structures and planning problems.   
\subsection{Observation data}
In the sphere environment, the observation data is a point-cloud converted into a voxel map of size $40 \times 40 \times 40$. However, for the other high-dimensional robot environments (Bartender and Kitchen), there exist workspace and entire scene observations at any time instant $t$. The workspace observation includes the current $\boldsymbol{x}_t \in \mathcal{X}$ and the target $\boldsymbol{x}_T \in \mathcal{X}$. The current workspace observation $\boldsymbol{x}_t$ at a given time is represented by each objects' poses and the robot end-effector pose. The target $\boldsymbol{x}_T$ is represented by the objects' target poses at the completion of the entire task. The scene observation is also a function of time represented as a voxel map $\boldsymbol{v}_t$ at instant $t$. We obtain raw point-cloud data from multiple Kinect sensors and process into voxel maps. The Kinect sensors are placed in the bartender and kitchen environments leading to voxel maps of dimensions $33\times 33\times 33$ and $32\times 32\times 32$, respectively.

\subsection{NTP2: Programs and API Arguments Set}
In our NTP2 setup, the list of initial programs includes $\mathrm{arrange\_table}$ and $\mathrm{swap\_tray\_objs}$. The Bartender (setup 1 and 2) and Kitchen tasks begin with the former, whereas the Bartender setup 3 begins with the latter program. The initial program can call either $\mathrm{pick\_place}$, $\mathrm{subgoal\_gen}$, $\mathrm{return\_arm}$, or $\mathrm{no\_op}$ programs followed by their underlying API-programs named $\mathrm{pick}$ and $\mathrm{place}$.  The API-programs $\mathrm{pick}$ and $\mathrm{place}$ represent an unconstrained planning problem, requiring a robot to reach a given target/object, and a constrained planning problem, demanding manipulation under manifold constraints, respectively.  An API-program also gets an argument, predicted by the API-decoder, which in our cases, is one of the objects (e.g., juice can, fuze bottle, soda can, etc.) to be picked or placed in the given scenario. Furthermore, the program $\mathrm{return\_arm}$ requires a robot to return to its initial default configuration from any starting state, and the program $\mathrm{no\_op}$ means no operation needed. Finally, the $\mathrm{subgoal\_gen}$ is executed to move objects acting as obstacles out of the way through pick-place procedures to achieve the desired sub-task.  

\begin{figure*}[t]
    \centering
    \begin{subfigure}[b]{0.3\textwidth}
       \includegraphics[width=5.1cm]{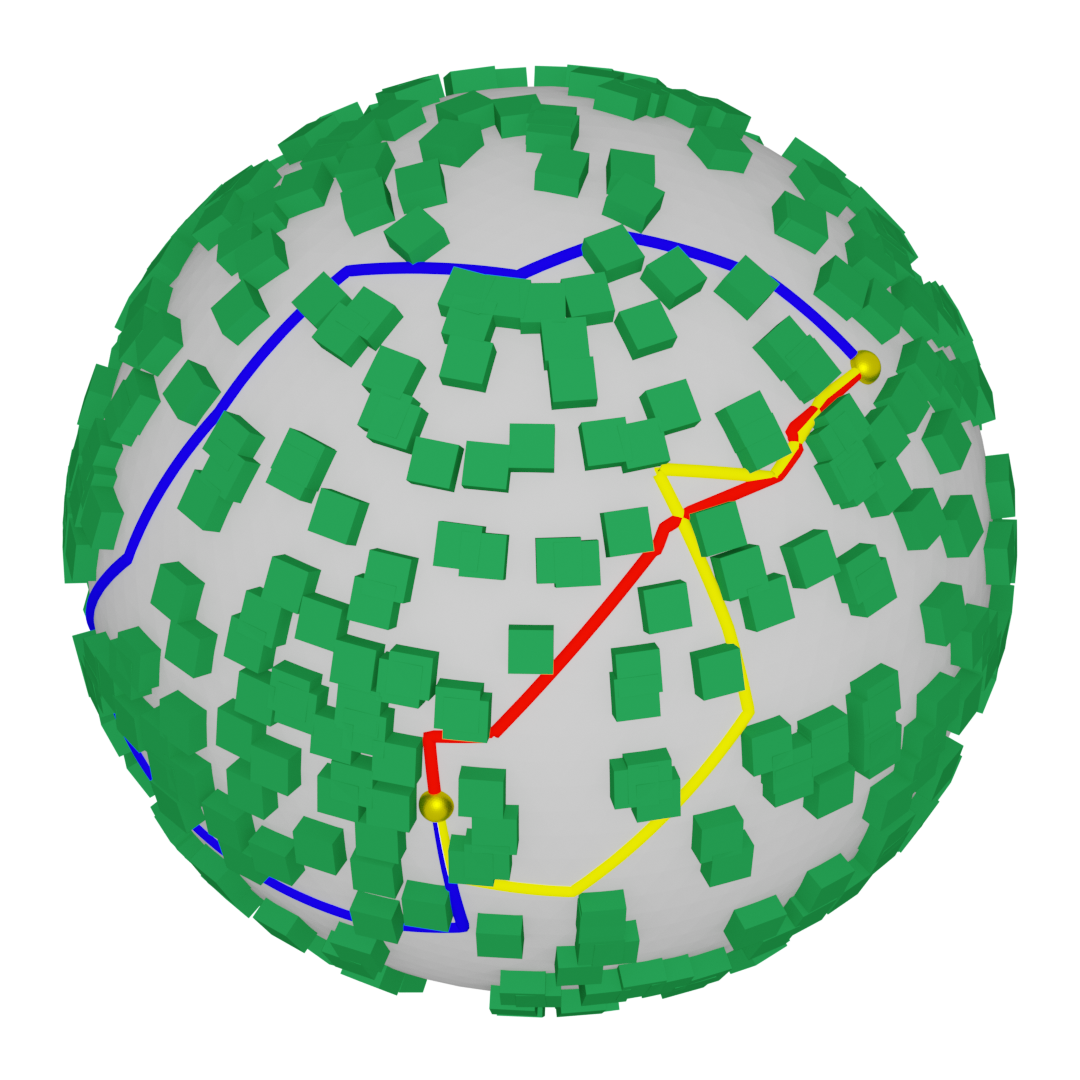}
       \caption{}
    \end{subfigure}    
    \begin{subfigure}[b]{0.3\textwidth}
     \includegraphics[width=5.1cm]{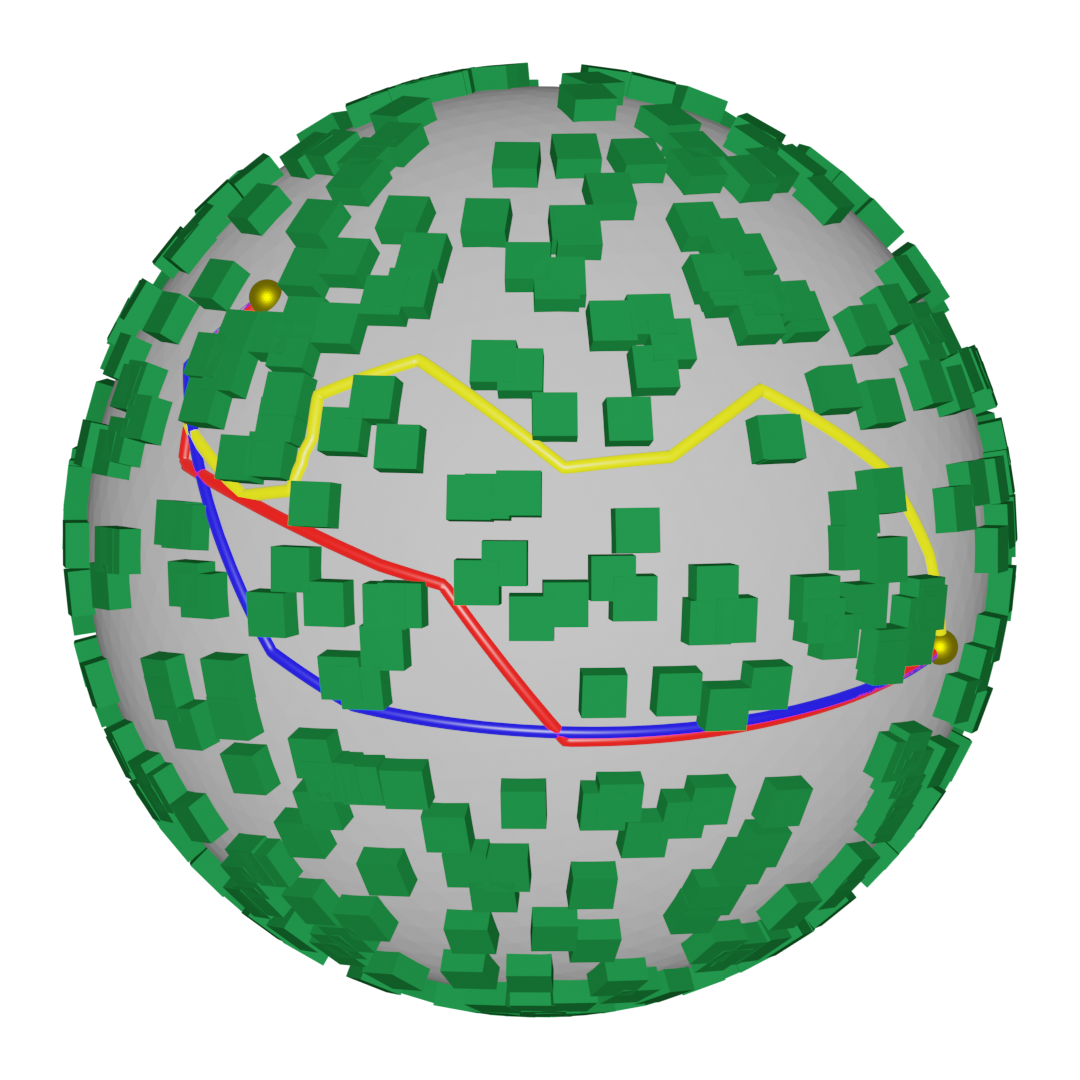}
     \caption{}
    \end{subfigure}
    \begin{subfigure}[b]{0.3\textwidth}
       \includegraphics[width=5.1cm]{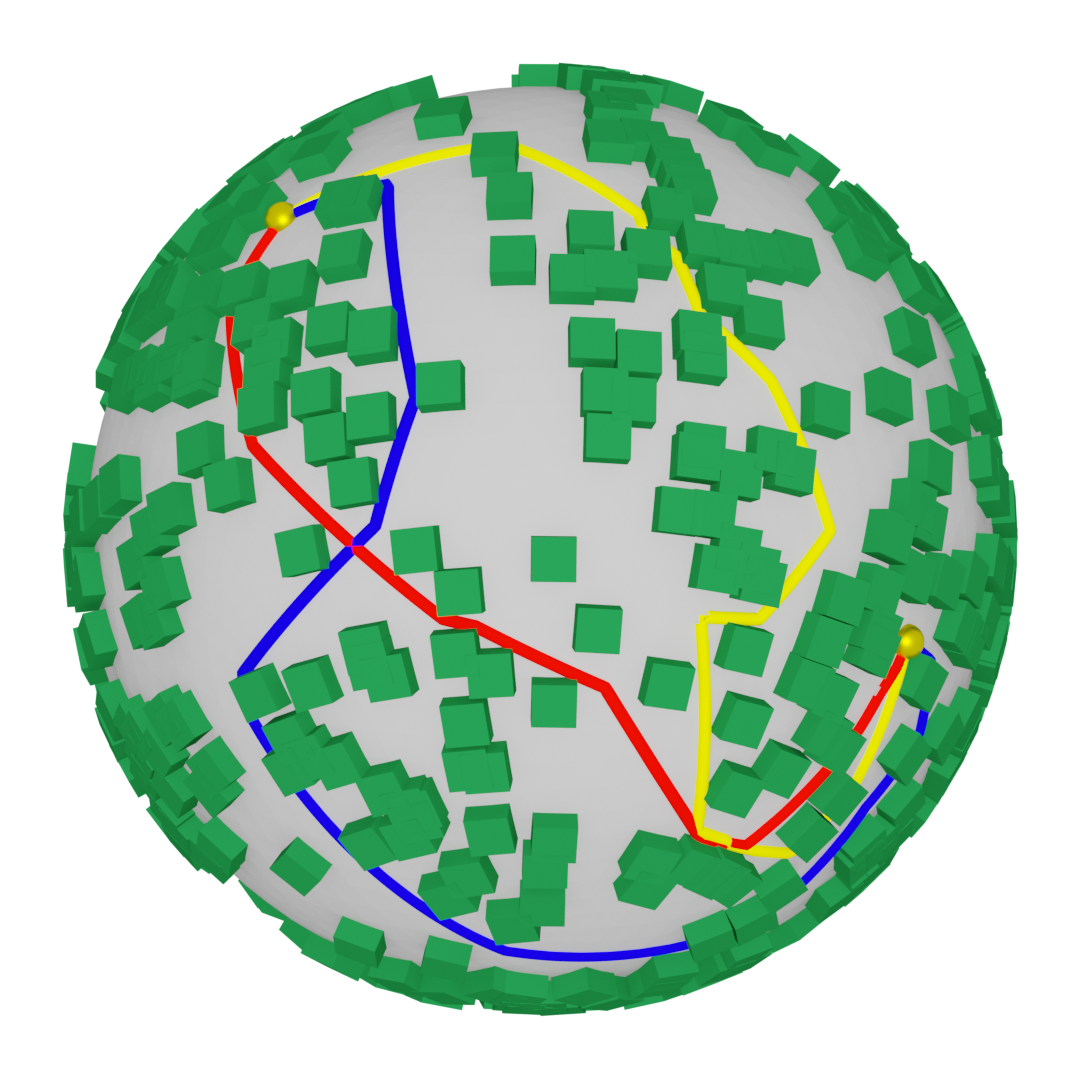}
       \caption{}
    \end{subfigure}
    \caption{Sphere Environment (Scenario 1): The paths found by CoMPNetX-FMT* (red), FMT* (yellow) and RRTConnect (blue) with atlas operator in three example scenes. It can be seen that CoMPNetX finds shorter, near-optimal paths compared to other methods.}\label{sphere-sc1}
\end{figure*}

\begin{figure*}[t]
    \centering
    \begin{subfigure}[b]{0.3\textwidth}
       \includegraphics[width=5.1cm]{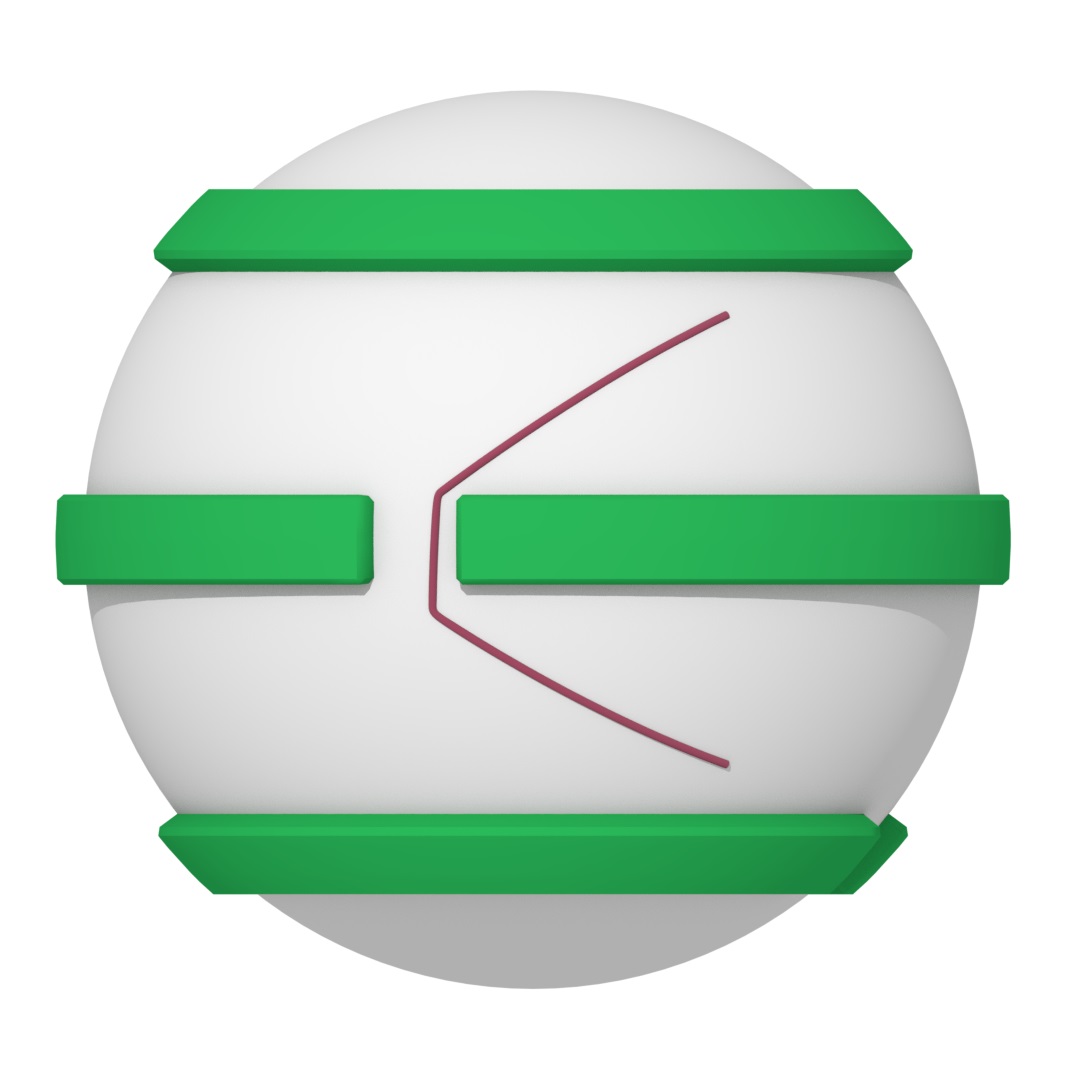}
       \caption{Planning Problem}
    \end{subfigure}    
    \begin{subfigure}[b]{0.3\textwidth}
     \includegraphics[width=5.1cm]{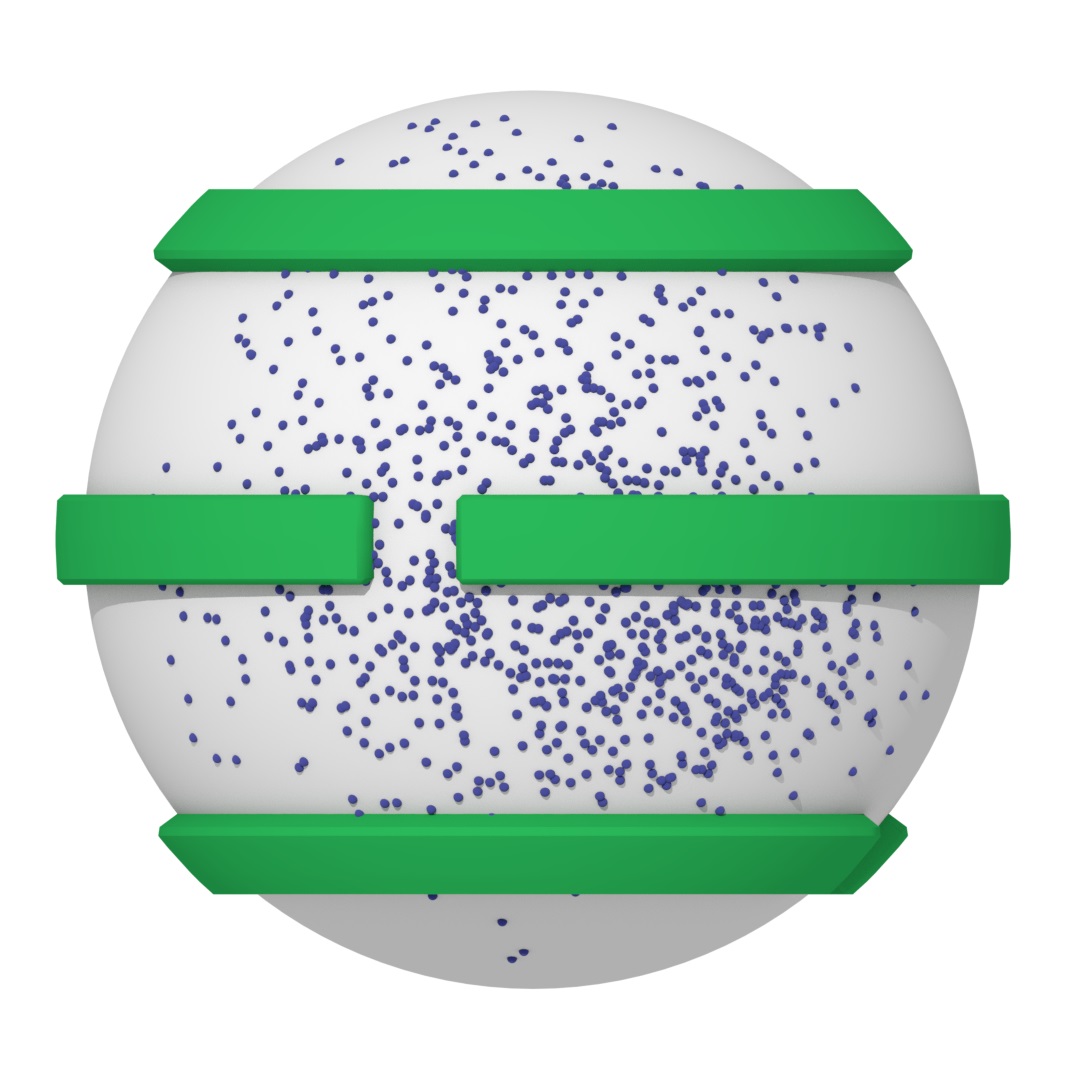}
     \caption{CoMPNetX Sampling}
    \end{subfigure}
    \begin{subfigure}[b]{0.3\textwidth}
       \includegraphics[width=5.1cm]{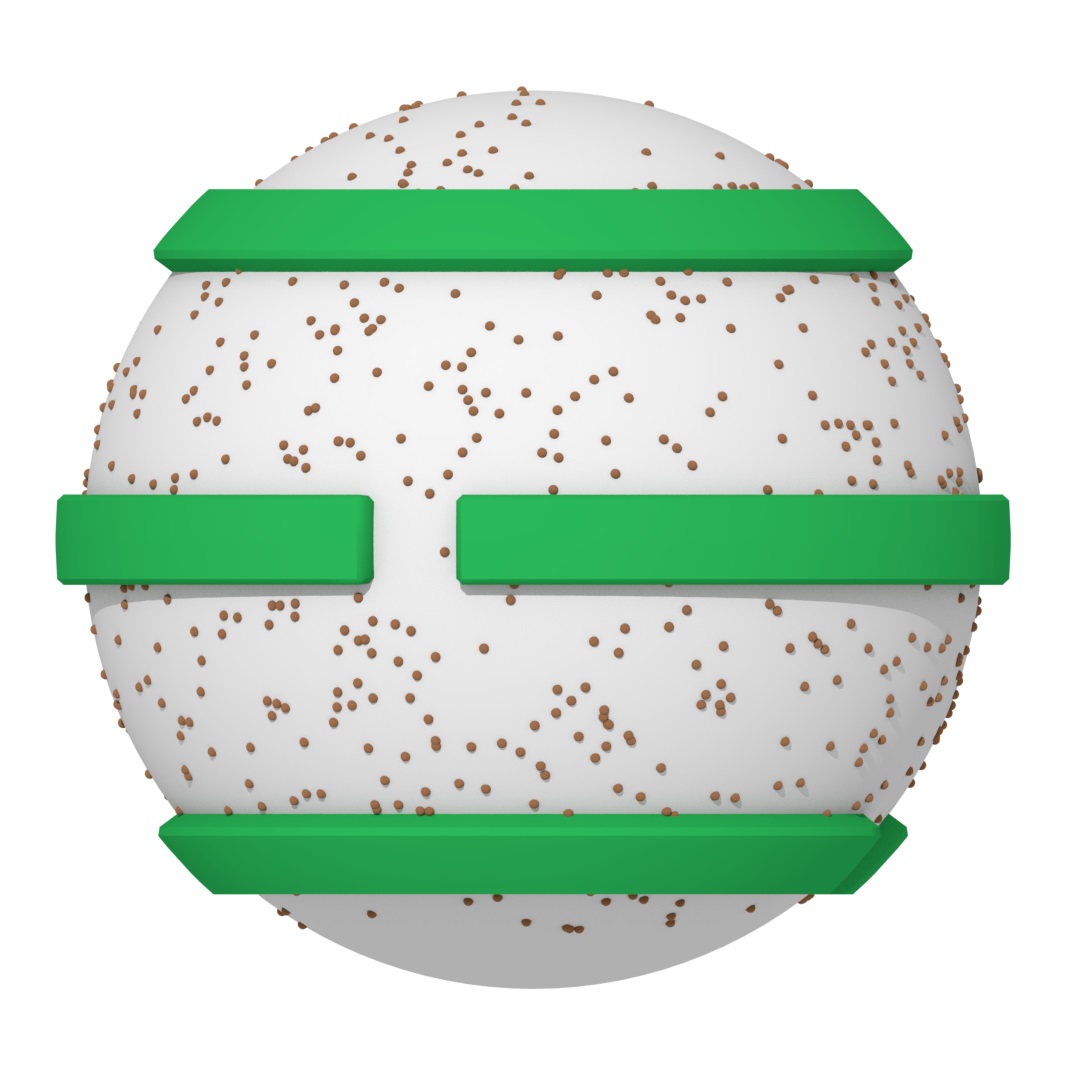}
       \caption{Uniform Sampling}
    \end{subfigure}
    \caption{Sphere Environment: CoMPNetX stochastically generates samples in the sub-space of a given region that potentially contains a path solution by leveraging past planning experiences and the environment perception information. It contrasts with traditional approaches that randomly explore the entire space and therefore struggle in high-dimensional planning problems to find a path solution.}\label{sphere2}
\end{figure*}
\begin{figure*}
    \centering
    \begin{subfigure}[b]{0.24\textwidth}
       \includegraphics[width=3.9cm]{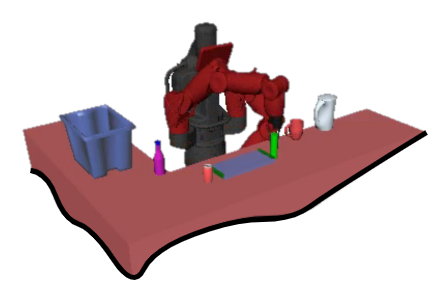}
    \end{subfigure}
    \begin{subfigure}[b]{0.24\textwidth}
       \includegraphics[width=3.9cm]{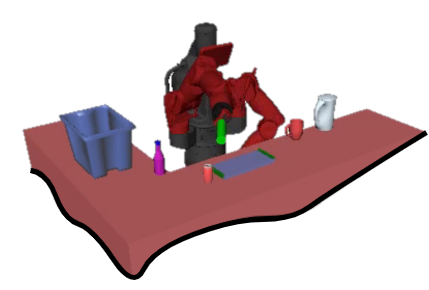}
    \end{subfigure}
        \begin{subfigure}[b]{0.24\textwidth}
       \includegraphics[width=3.9cm]{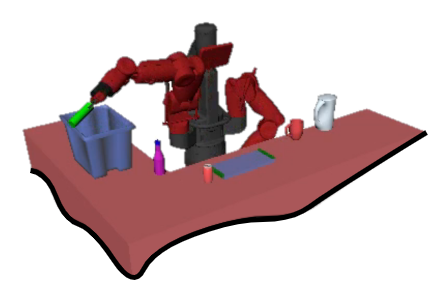}
    \end{subfigure}
    \begin{subfigure}[b]{0.24\textwidth}
       \includegraphics[width=3.9cm]{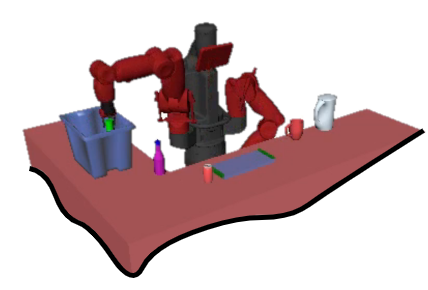}
    \end{subfigure}\vspace*{0.05in}
    \caption*{(a) Move juice can to the trash bin.} \vspace*{0.05in}
    \begin{subfigure}[b]{0.24\textwidth}
       \includegraphics[width=3.9cm]{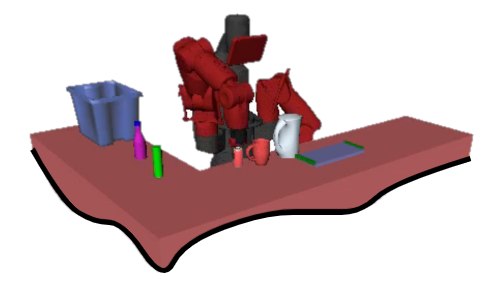}
    \end{subfigure}
    \begin{subfigure}[b]{0.24\textwidth}
       \includegraphics[width=3.9cm]{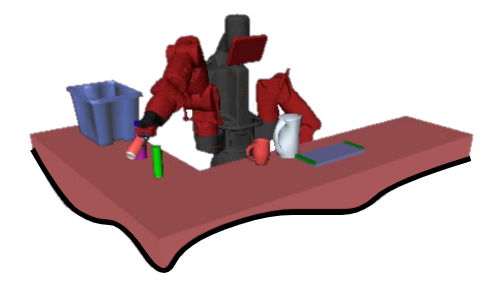}
    \end{subfigure}
        \begin{subfigure}[b]{0.24\textwidth}
       \includegraphics[width=3.9cm]{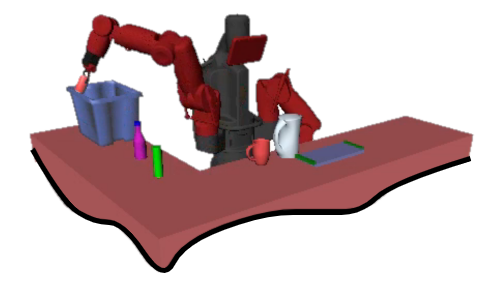}
    \end{subfigure}
    \begin{subfigure}[b]{0.24\textwidth}
       \includegraphics[width=3.9cm]{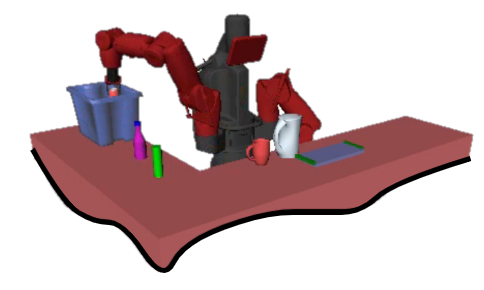}
    \end{subfigure}\vspace*{0.05in}
    \caption*{(b) Move soda can to the trash bin.}\vspace*{0.05in}
      \begin{subfigure}[b]{0.24\textwidth}
       \includegraphics[width=3.9cm]{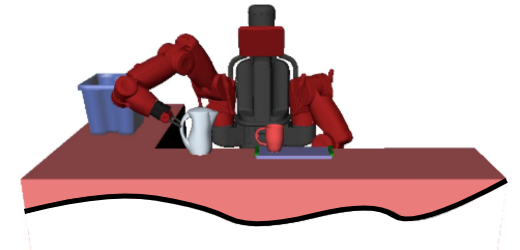}
    \end{subfigure}
    \begin{subfigure}[b]{0.24\textwidth}
       \includegraphics[width=3.9cm]{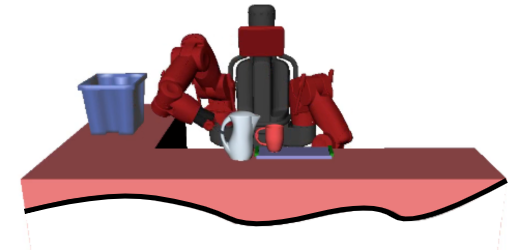}
    \end{subfigure}
        \begin{subfigure}[b]{0.24\textwidth}
       \includegraphics[width=3.9cm]{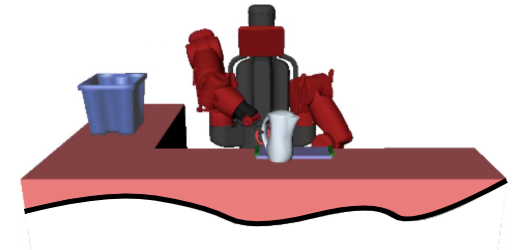}
    \end{subfigure}
    \begin{subfigure}[b]{0.24\textwidth}
       \includegraphics[width=3.9cm]{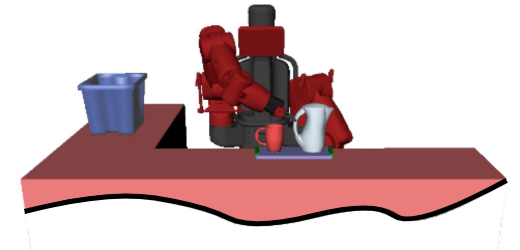}
    \end{subfigure}\vspace*{0.05in}
    \caption*{(c) Carefully place the kettle, i.e., without tilting, onto the tray.}\vspace*{0.05in}
    \caption{Bartender setup (SC1): Figs. (a-c) show CoMPNetX motion sequences of moving juice can, soda can, and kettle to their targets in three different test cases.}\label{bartender}\end{figure*}
  
\begin{figure*}[t]
    \centering
    \begin{subfigure}[b]{0.24\textwidth}
       \includegraphics[width=3.9cm]{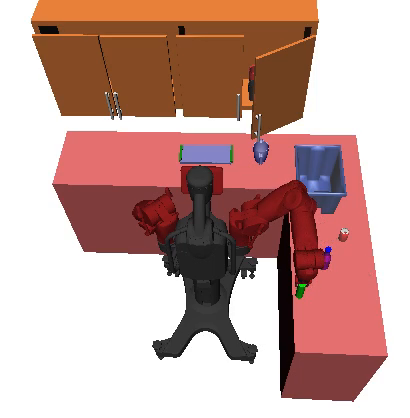}
    \end{subfigure}
    \begin{subfigure}[b]{0.24\textwidth}
       \includegraphics[width=3.9cm]{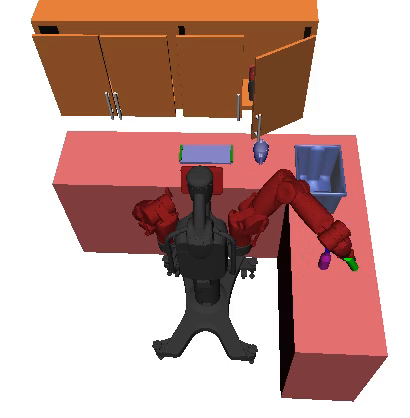}
    \end{subfigure}
        \begin{subfigure}[b]{0.24\textwidth}
       \includegraphics[width=3.9cm]{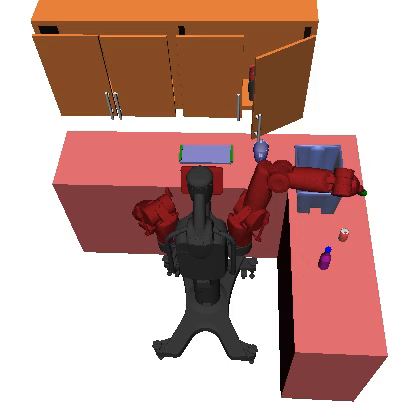}
    \end{subfigure}
    \begin{subfigure}[b]{0.24\textwidth}
       \includegraphics[width=3.9cm]{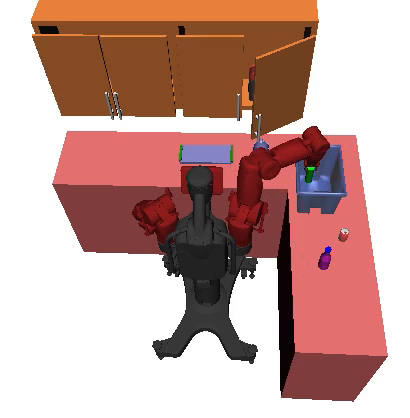}
    \end{subfigure}\vspace*{0.05in}
    \caption*{(a) Move the juice can to the trash.}\vspace*{0.1in}
    \begin{subfigure}[b]{0.24\textwidth}
       \includegraphics[width=3.9cm]{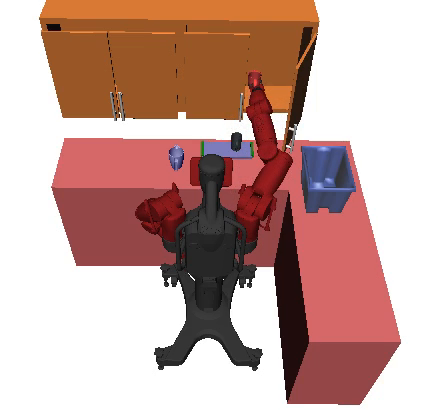}
    \end{subfigure}
    \begin{subfigure}[b]{0.24\textwidth}
       \includegraphics[width=3.9cm]{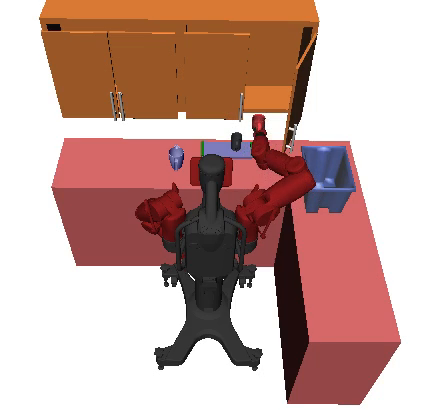}
    \end{subfigure}
        \begin{subfigure}[b]{0.24\textwidth}
       \includegraphics[width=3.9cm]{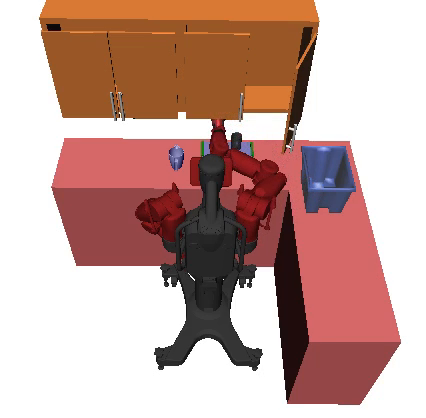}
    \end{subfigure}
    \begin{subfigure}[b]{0.24\textwidth}
       \includegraphics[width=3.9cm]{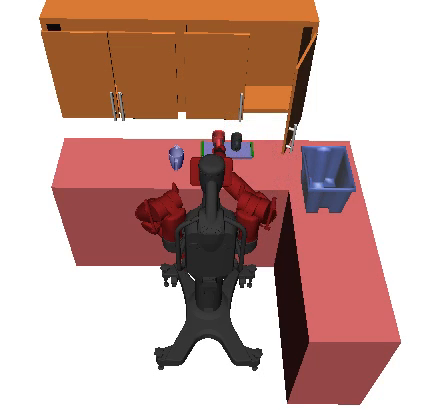}
    \end{subfigure}\vspace*{0.05in}
    \caption*{(b) Carefully place, without tilting, the red mug from cabinet onto the tray.}\vspace*{0.1in}
      \begin{subfigure}[b]{0.24\textwidth}
       \includegraphics[width=3.9cm]{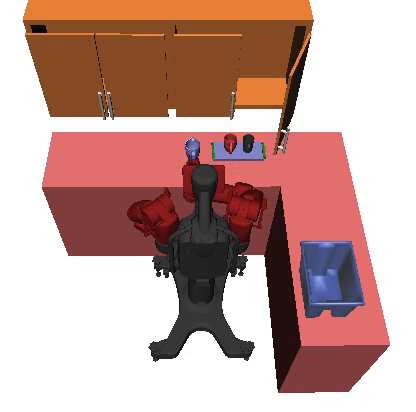}
    \end{subfigure}
    \begin{subfigure}[b]{0.24\textwidth}
       \includegraphics[width=3.9cm]{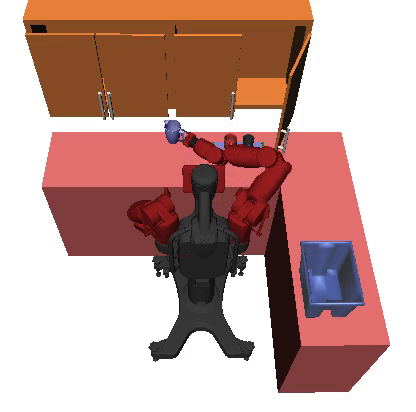}
    \end{subfigure}
        \begin{subfigure}[b]{0.24\textwidth}
       \includegraphics[width=3.9cm]{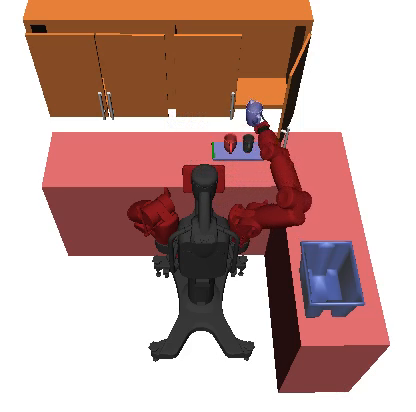}
    \end{subfigure}
    \begin{subfigure}[b]{0.24\textwidth}
       \includegraphics[width=3.9cm]{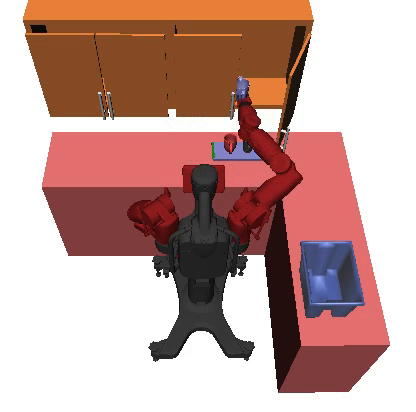}
    \end{subfigure}\vspace*{0.05in}
    \caption*{(c) Carefully move pitcher into the cabinet.}\vspace*{0.05in}
\caption{Kitchen setup: Figs. (a-c) show instances of CoMPNetX planned motions for moving the juice can, red mug, and pitcher to their targets under various constraints in three different test scenarios.}\label{kitchen}    
\end{figure*}

\begin{figure*}
    \centering
       \includegraphics[width=17.2cm]{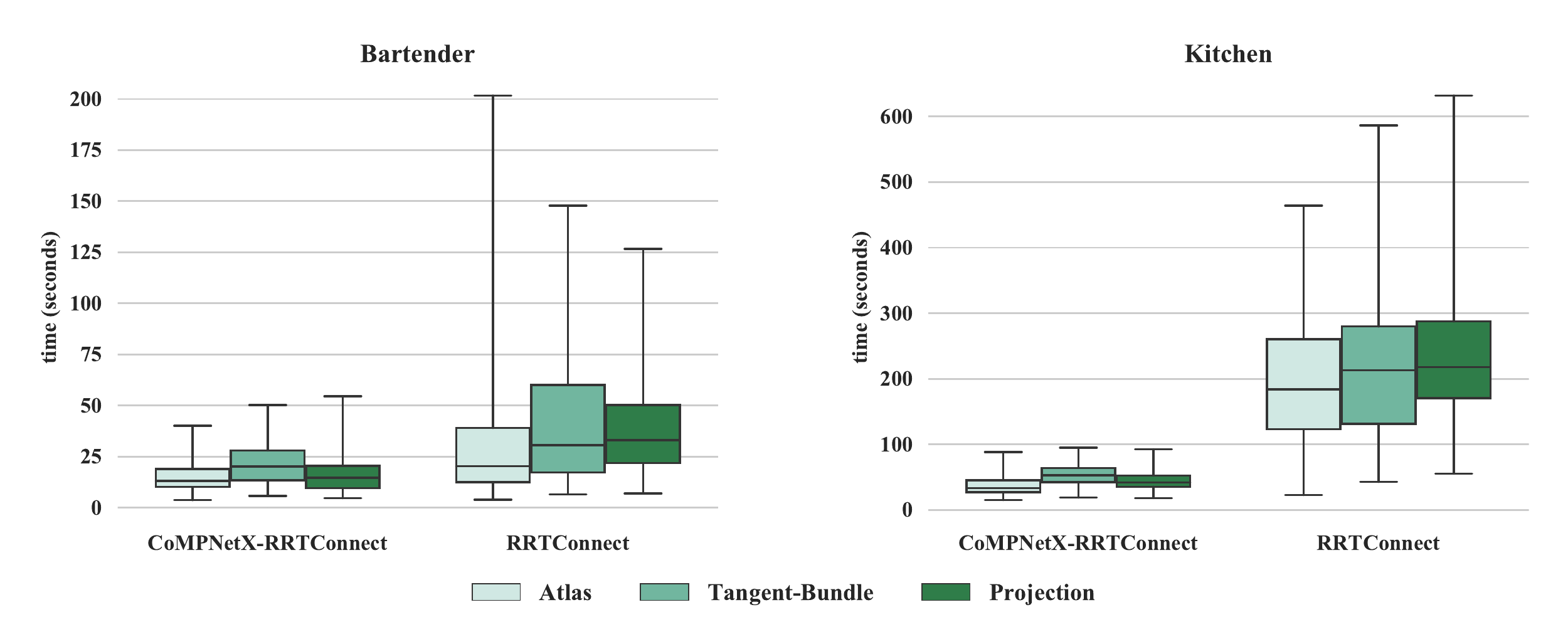} \caption{The boxplots show the total computation times of CoMPNetX-RRTConnect and traditional RRTConnect with atlas, tangent-bundle, and projection-based constraint adherence approaches, solving all constrained manipulation problems in the bartender and kitchen environments. It can be seen that CoMPNetX is significantly faster with significantly lower computation times and shows lower variability than classical RRTConnect (equivalent to Atlas-RRT \cite{jaillet2017path}, TB-RRT \cite{kim2016tangent}, and CBiRRT \cite{berenson2011task}). 
}\label{boxplot}
\end{figure*}

\section{Results}
In this section, we present the results and analysis of the following evaluation studies: (i) A comparison study evaluating CoMPNetX and state-of-the-art classical SMP planning methods with an underlying constraint-adherence approach (projection, atlas, or tangent bundle) on unseen challenging problems in environments named Sphere, Bartender, and Kitchen. (ii) An ablation study comparing CoMPNetX with its ablated models and our previous method CoMPNet~\cite{qureshi2020compnet}. (iii) An extended evaluation to highlight the mutualistic relationship of learning-based task programmers and CoMPNetX and their capacity to generalize across different planning domains.


\subsection{Comparative analysis}
This section compares SMP methods augmented with batch and bidirectional CoMPNetX against their classical setups in solving CMP problems. In the batch method, we select FMT* \cite{janson2015fast}, a state-of-the-art classical SMP algorithm, and it is proven to perform better than standard approaches like RRT* and PRM* \cite{karaman2011sampling}. FMT* begins with an initial batch of $N_{init}$ uniform samples, including a goal configuration. In case, the initial set of samples does not yield a path solution, FMT* continues to expand the tree by generating a new random sample in every planning iteration. We choose FMT* to highlight the flexibility offered by CoMPNetX in generating sample batches with different $K\geq1$ according to the given SMP method. In CoMPNetX-FMT*, we generate an initial batch of $N_{init}$ samples with $K<<N_{init}$. The initial $K$ configurations are randomly sampled from an obstacle-free space, to form $K_{\boldsymbol{q}_{curr}}$, which are passed to CoMPNetX to obtain the next set of configurations $K_{\boldsymbol{q}_{next}}$. In the next step, the $K_{\boldsymbol{q}_{next}}$ becomes $K_{\boldsymbol{q}_{curr}}$, and the process of sample generation with CoMPNetX is repeated until it gathers an initial batch of size $N_{init}$, which also includes a goal state. Furthermore, in case, the initial batch does not yield a path solution, in every subsequent planning step, we randomly select a node in the FMT* tree as an input to CoMPNetX to generate a new informed sample towards the given target (see Algorithm \ref{algo:btree}). In the bidirectional CoMPNetX, we merge CoMPNetX into RRTConnect, as reported in Algorithm \ref{algo:bitree}.

In the sphere environment, we evaluate both batch and bidirectional CoMPNetX. In the first scenario (Fig. \ref{sphere-sc1}), CoMPNetX-FMT* with atlas, tangent-bundle, and projection exhibit similar performances with over $99\%$ success rates, computation times of about $0.89\pm 0.13$ seconds, and the path lengths of about $2.18\pm 0.014$ units. In contrast, FMT* with atlas, tangent-bundle, and projection appeared to have $1.22\pm 0.034$, $1.34\pm 0.061$, and $1.14\pm 0.027$ seconds computation times, around $95\%$, $89\%$ and $98\%$ success rates, respectively, and somewhat similar path lengths as CoMPNetX. Overall, COMPNetX-FMT* finds near-optimal paths with lower computation times and higher success rates than classical FMT*. On the other hand, CoMPNetX-RRTConnect and RRTConnect exhibits similar computation times of about $0.01$-$0.02$ seconds. However, the former's path lengths are significantly better than the latter approach, as also shown in Fig. \ref{sphere-sc1}.

The second sphere scenario, is entirely an unseen environment for CoMPNetX as it was only trained on scenario 1. In this case, CoMPNetX-FMT* with all types of constraint-adherence operators demonstrate performances with around $96\%$ success rates, computation times of about $0.55\pm 0.11$ seconds, and the path lengths of about $3.13\pm 1.8$ units. The classical FMT* with atlas, tangent-bundle, and projection takes $1.93\pm 1.34$, $2.05\pm 1.44$, and $0.95\pm 0.34$ seconds computation times with success rate of around $89\%$, $87\%$ and $100\%$, respectively. Generally, COMPNetX-FMT* found better quality paths, in terms of path lengths, with lower computation times and generalized to this new environment with high success rates comparable to classical FMT*. CoMPNetX-RRTConnect and RRTConnect performed similarly in terms of computation times but the latter provides poor quality path solutions. Fig. \ref{sphere2} depicts the exploration by CoMPNetX and uniform sampling for a given problem. It can be seen that our method explores the space that potentially contains a path solution, thus leading to better performance, and this becomes even more significant in high-dimensional problems, as presented in the remainder of this section.

In the Bartender (scenario 1) and Kitchen environments, to solve constrained manipulation tasks, we focus on bidirectional SMP methods only since they have become a standard tool for solving high-dimensional CMP problems, and other methods such as unidirectional SMP algorithms struggle in such cases and exhibit high computational times with low success rates~\cite{jaillet2017path, kim2016tangent, berenson2011task}. We evaluate CoMPNetX-RRTConnect and traditional RRTConnect with Atlas, Tangent Bundle, and Projection-based constrained-adherence methods, resulting in Atlas-RRT~\cite{jaillet2017path}, TB-RRT~\cite{kim2016tangent}, and CBiRRT~\cite{berenson2011task} algorithms. Figs. \ref{bartender} (a-c) show instances of CoMPNetX-RRTConnect generating motions in the Bartender (Scenario 1) for manipulating juice can (Fig. (a)), soda can (Fig. (b)), and kettle (Fig. (c)) in three different test scenarios. Likewise, Figs. \ref{kitchen} (a-c) shows CoMPNetX-RRTConnect motion sequences for moving juice can, red mug, and pitcher in three different kitchen scenarios. Furthermore, Fig. \ref{compnet_exe} displayed waypoints generated by our approach for the cabinet's door opening task.

In these high-dimensional CMP scenarios, Fig. \ref{boxplot} provides the box plots of the total computational time to solve all the manipulation tasks. Table \ref{tab1} presents the mean success rates with their standard deviations of all methods. Furthermore, Table \ref{tab2} compares the mean computation times with standard deviations for the individual objects, grouped by their constraint types, in each of the scenarios. 

It can be seen that our method exhibits significantly lower inter-quartile computational time ranges with a narrow spread than other methods while retaining similar success rates. Moreover, the results also show that with the increasing complexity of the planning problems from sphere to kitchen environment, the computation times of traditional methods increase significantly with large standard deviations compared to our approach. For instance, between bartender and kitchen task (Table \ref{tab2}), the planning times of manipulation under collision-avoidance and stability constraints increase from $\sim1$ to $\sim7$ seconds for CoMPNetX-RRTConnect and from $\sim1$ to $\sim40$ seconds for traditional RRTConnect. Furthermore, in these experiments, we randomize the positioning of objects in the environments, so the models are shown to generalize to new,  unseen objects' positioning. However, these models can also generalize to new objects if trained accordingly with a variety of different items.
\begin{table}
    \centering 
    \begin{tabular}{cccc}
        \toprule
        \multirow{2}{*}{Algorithms}
            &\multirow{1}{*}{Type of}
            &\multirow{1}{*}{Bartender}
            &\multirow{1}{*}{Kitchen}\\
            &\multirow{1}{*}{Constraint-Adherence}
            &\multirow{1}{*}{$(\%)$}
            &\multirow{1}{*}{$(\%)$}\\
        \midrule
        \multirow{3}{*}{RRTConnect}
            &\multirow{1}{*}{Projection}
                &\multirow{1}{*}{$98.3 \pm 1.1$} 
                &\multirow{1}{*}{$87.7 \pm 4.8$} 
                \\
            &\multirow{1}{*}{Atlas}
                &\multirow{1}{*}{$99.2 \pm 0.6$} 
                &\multirow{1}{*}{$95.4 \pm 3.8$} \\
            &\multirow{1}{*}{Tangent-bundle}
                &\multirow{1}{*}{$97.8 \pm 2.4$} 
                &\multirow{1}{*}{$90.1 \pm 5.7$}  \\ 
        \midrule
        
            \multirow{2}{*}{CoMPNetX-}&\multirow{1}{*}{Projection}
                &\multirow{1}{*}{$98.3 \pm 0.6$} 
                &\multirow{1}{*}{$88.4 \pm 1.8$} \\
            &\multirow{1}{*}{Atlas}
                &\multirow{1}{*}{$99.8 \pm 0.1$} 
                &\multirow{1}{*}{$95.3 \pm 1.3$}\\
            \multirow{1}{*}{RRTConnect}&\multirow{1}{*}{Tangent-bundle}
                &\multirow{1}{*}{$97.8 \pm 2.3$} 
                &\multirow{1}{*}{$90.4 \pm 2.7$}\\ 
        \bottomrule
    \end{tabular}
    \caption{The total mean success rates with standard deviations, over five trials, of CoMPNetX-RRTConnect and traditional RRTConnect for solving all manipulation problems in the bartender and kitchen environments.} \label{tab1}
\end{table}    

\begin{table*}[t]
    \centering 
    \begin{tabular}{ccccccc}
        \toprule
        \multirow{2}{*}{Algorithms}
            &\multirow{1}{*}{Type of}
            &\multicolumn{2}{c}{Bartender}
            &\multicolumn{3}{c}{Kitchen}\\
            \cmidrule(lr){3-4}\cmidrule(lr){5-7}
            &\multirow{1}{*}{Constraint-Adherence}
            &\multicolumn{1}{c}{J/F/S}
            &\multicolumn{1}{c}{R/K}
            &\multicolumn{1}{c}{J/F/S}
            &\multicolumn{1}{c}{C}
            &\multicolumn{1}{c}{R/B/P}\\
        \midrule
        \multirow{3}{*}{RRTConnect}
            &\multirow{1}{*}{Projection}
                &\multirow{1}{*}{$12.64 \pm 8.21$} 
                &\multirow{1}{*}{$1.06 \pm 0.87$} 
                &\multirow{1}{*}{$32.64 \pm 22.40$}
                &\multirow{1}{*}{$0.05 \pm 0.04$}
                &\multirow{1}{*}{$49.79 \pm 22.96$}\\
            &\multirow{1}{*}{Atlas}
                &\multirow{1}{*}{$10.86 \pm 11.08$} 
                &\multirow{1}{*}{$0.93 \pm 0.82$}  
                &\multirow{1}{*}{$24.87 \pm 19.81$}
                &\multirow{1}{*}{$0.04 \pm 0.03$}
                &\multirow{1}{*}{$41.28 \pm 24.02$}\\
            &\multirow{1}{*}{Tangent-bundle}
                &\multirow{1}{*}{$16.23 \pm 14.78$} 
                &\multirow{1}{*}{$1.68 \pm 0.82$}  
                &\multirow{1}{*}{$27.54 \pm 21.47$}
                &\multirow{1}{*}{$0.05 \pm 0.03$}
                &\multirow{1}{*}{$46.61 \pm 26.06$}\\ 
        \midrule
        \multirow{3}{*}{CoMPNetX-RRTConnect}
            &\multirow{1}{*}{Projection}
                &\multirow{1}{*}{$4.59 \pm 2.75$} 
                &\multirow{1}{*}{$1.17 \pm 1.21$}  
                &\multirow{1}{*}{$8.02 \pm 3.34$}
                &\multirow{1}{*}{$0.04 \pm 0.01$}
                &\multirow{1}{*}{$7.70 \pm 3.63$}\\
            &\multirow{1}{*}{Atlas}
                &\multirow{1}{*}{$4.51 \pm 2.24$} 
                &\multirow{1}{*}{$0.77 \pm 0.38$}  
                &\multirow{1}{*}{$6.26 \pm 3.44$}
                &\multirow{1}{*}{$0.02 \pm 0.01$}
                &\multirow{1}{*}{$5.84 \pm 2.78$}\\
            &\multirow{1}{*}{Tangent-bundle}
                &\multirow{1}{*}{$6.32 \pm 3.14$} 
                &\multirow{1}{*}{$1.21 \pm 1.09$}  
                &\multirow{1}{*}{$8.68 \pm 3.34$}
                &\multirow{1}{*}{$0.04 \pm 0.02$}
                &\multirow{1}{*}{$9.62 \pm 3.37$}\\ 
        \bottomrule
    \end{tabular}
    \caption{The mean computation times with standard deviations of CoMPNetX-RRTConnect and classical RRTConnect algorithm with underlying projection, atlas, and tangent-bundle integerators in solving the manipulation planning problems for the individual objects. The objects are denoted by their first letter and are grouped by their constraint types. It can be seen that CoMPNetX computation times are not only lower but also more consistent across different problems than other approaches.} \label{tab2}
\end{table*}

\begin{table*}[b]
\centering 
{
\begin{tabular}{ccccc}\toprule
\multirow{2}{*}{Tasks}&\multicolumn{4}{c}{Algorithms with Atlas Integrator and an underlying RRTConnect}\\\cmidrule{2-5}
&\multicolumn{1}{c}{CoMPNetX}&\multicolumn{1}{c}{CoMPNetX (w/o NProj)}&\multicolumn{1}{c}{CoMPNet}&\multicolumn{1}{c}{Continuation-based Sampling}\\\midrule


\multirow{1}{*}{Bartender}& \multirow{1}{*}{$\boldsymbol{15.05 \pm 06.86}$ $\boldsymbol{(99.8\%)}$} & \multirow{1}{*}{$17.31 \pm 09.10$ $(98.1\%)$}  &\multirow{1}{*}{$19.77 \pm 10.84$ $(94.3\%)$}&\multirow{1}{*}{$33.34 \pm 33.16$ $(99.2\%)$}\\ 

\multirow{1}{*}{Kitchen}& \multirow{1}{*}{$\boldsymbol{36.47 \pm 14.16}$ $\boldsymbol{(95.4\%)}$} & \multirow{1}{*}{$42.57 \pm 14.94$ $(93.0\%)$}  &\multirow{1}{*}{ $46.93 \pm 15.79$ $(90.8\%)$}&\multirow{1}{*}{$194.17 \pm 96.39$ $(95.4\%)$}\\ \bottomrule
\end{tabular}}
\caption{The total mean computation times with standard deviations and mean success rates are presented for various sampling approaches with an underlying RRTConnect and atlas-based integrator. The sampling approaches include: i) CoMPNetX, ii) CoMPNetX without neural gradient-based projects ($\mathrm{NProj}$), iii) previously proposed CoMPNet, i.e., with text-based task specifications and without $\mathrm{NProj}$, and iv) the traditional conitnuation-based sampling, in the bartender and kitchen environments.} \label{tab3}
\vspace*{-0.1in}\end{table*}

\subsection{Ablative analysis}
In this analysis, we ablate various components of CoMPNetX to highlight their significance in solving complex CMP problems. Table \ref{tab3} summarizes the results with mean total computation time and their standard deviation, and mean success rates for solving all manipulation problems in the Bartender (scenario 1) and Kitchen environments. 

The first alteration is to remove the $\mathrm{NProj}$, i.e., the neural discriminator's gradient-based projections (Eqn. (9)), from CoMPNetX. Note that in Algorithm 5, we use $\mathrm{NProj}$ only when the distance of generated configurations from the manifold is greater than $\nu$. The value of $\nu$ is selected to be a positive scalar multiple of the tolerance $\epsilon$ so that to fix those configurations that are an order of magnitude distance away from the manifold than the allowed tolerance. It can be seen that there are only a fraction of cases (2-3 $\%$) where generated configurations by the Neural Generator were not close to the manifold in constrained planning problems and fixing them with $\mathrm{NProj}$ led to performance gains.

The second ablation is to evaluate the impact of neural task representations on CoMPNetX. In our previous work \cite{qureshi2020compnet}, we show that task-representations are crucial for CoMPNet. In that setting, text-based task representations led to significant improvements in performance than CoMPNet without any task-representations. Moreover, the results also highlighted that text-based representations become better than simple one-hot encoding when the number of tasks increases, e.g., from bartender to kitchen environments. It is because one-hot representations become very limited in practice with a growing set of multi-task and multimodal constraints. In this study, we now compare the neural task and text-based task representations for constrained neural motion planning. Table \ref{tab3} presents the comparison of CoMPNetX (with neural task representations and without $\mathrm{NProj}$) and CoMPNet~\cite{qureshi2020compnet} (with text-based task representation and without $\mathrm{NProj}$) in solving all the manipulation problems in the bartender and kitchen environments. The results indicate that the former, i.e., neural task representations, leads to better performance than the latter in computation times and success rates. Moreover, the statistical paired testing of these two methods resulted in p-values of $1.13\times 10^{-06}$ and $4.84\times 10^{-07}$ in the bartender and kitchen environments, which validates that CoMPNetX outperforms CoMPNet~\cite{qureshi2020compnet} by a significant margin. The reason is that the neural task representations consider the workspace observation and the overall program hierarchy, whereas the text-representations are agnostic of underlying task semantics. Furthermore, the learning-based task programmer not just provide task representations but also generate a task plan that saves lot of effort in hand-engineering sub-task sequences. Nevertheless, despite all ablations of CoMPNetX, it can be seen that our method performs significantly better than classical sampling techniques.

\subsection{Extended Analysis: Mutual Symbiotic Relationship}
In our comparative analysis, we show that CoMPNetX generalizes to new locations of the objects (i.e., not seen during training) and solves those practical problems in few seconds where gold standard SMP methods take up to several minutes to obtain comparable success rates. In this extended analysis, we show the joint operation of learning-based task programmer and CoMPNetX and evaluate our models, trained on bartender (scenario 1), for further generalization to new problems, such as in bartender scenarios 2 and 3, to simultaneously solve both unconstrained and constrained planning problems. Note that our trained model on bartender scenario 1 never had cases where either start, goal, or both states of the given sub-task object were occupied by other objects, acting as obstacles. Therefore, the planner needs to move them out of the way before accomplishing the desired sub-task. 

Table \ref{tab4} presents the total mean computation times with mean success rates of CoMPNetX for solving all unconstrained ($\mathrm{pick}$) and constrained ($\mathrm{place}$) tasks in the bartender scenarios 2 and 3. In these scenarios, the NTP2 success rate was about $95\%$ and $89\%$, and from those successful cases, CoMPNetX achieves around $90\%$ and $80\%$ success rate, respectively, in solving given motion planning problems. In unconstrained planning problems, CoMPNetX calls an underlying MPNet algorithm \cite{qureshi2019motionb}, solving problems in 2-3 seconds computation time, i.e., the computational gains over gold-standard SMPs are retained for unconstrained problems as well. In constrained planning problems, CoMPNetX uses RRTConnect, and their computation times for individual tasks were similar to reported for Bartender scenario 1.

\begin{figure*}
    \centering
       \includegraphics[width=17.0cm]{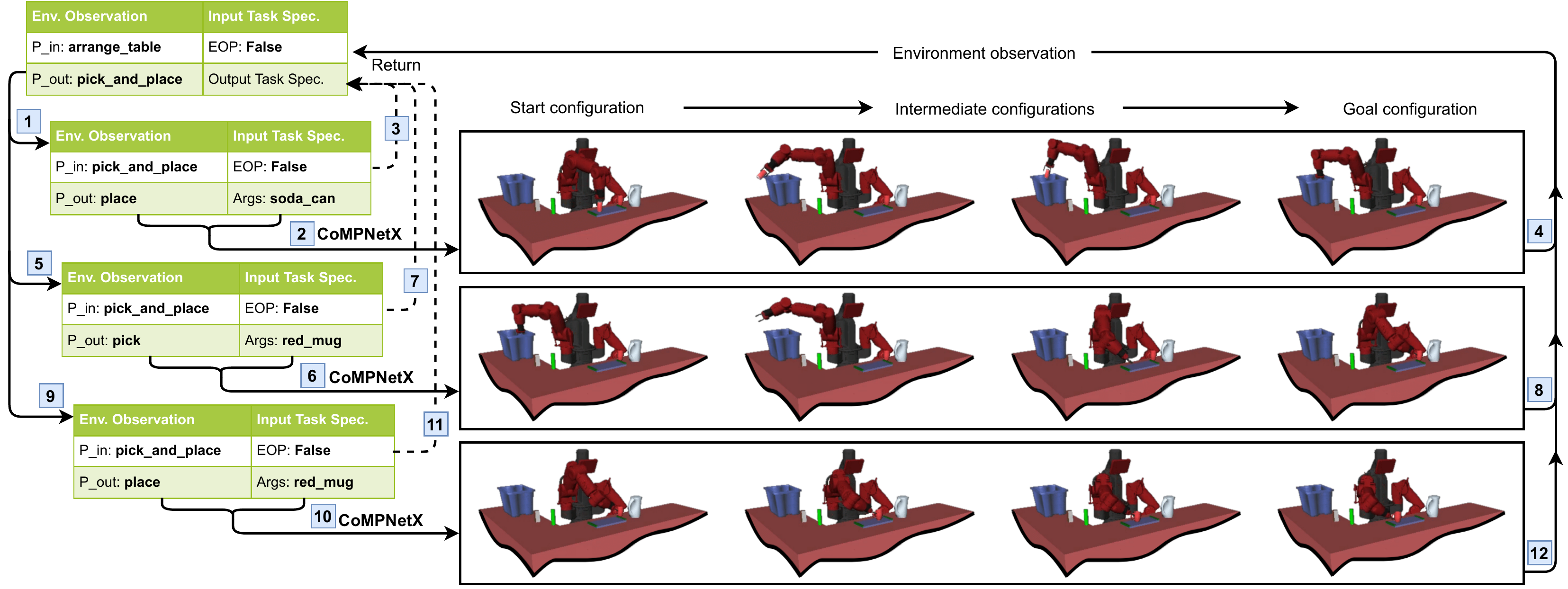}\vspace*{-0.0in}
    \caption{A mutual symbiotic operation of a learning-based task programmer, i.e., NTP2 and CoMPNetX for solving both unconstrained and constrained planning problems in the Bartender scenario 2. The numbers in small boxes indicate the order in which the procedures are executed. The task programmer generates a sub-task and their representation using which the CoMPNetX accomplishes them by outputting a feasible robot motion for interaction with the environment. 
}\label{bar_sc2}
\end{figure*}

\begin{figure*}
    \centering
       \includegraphics[width=17.0cm]{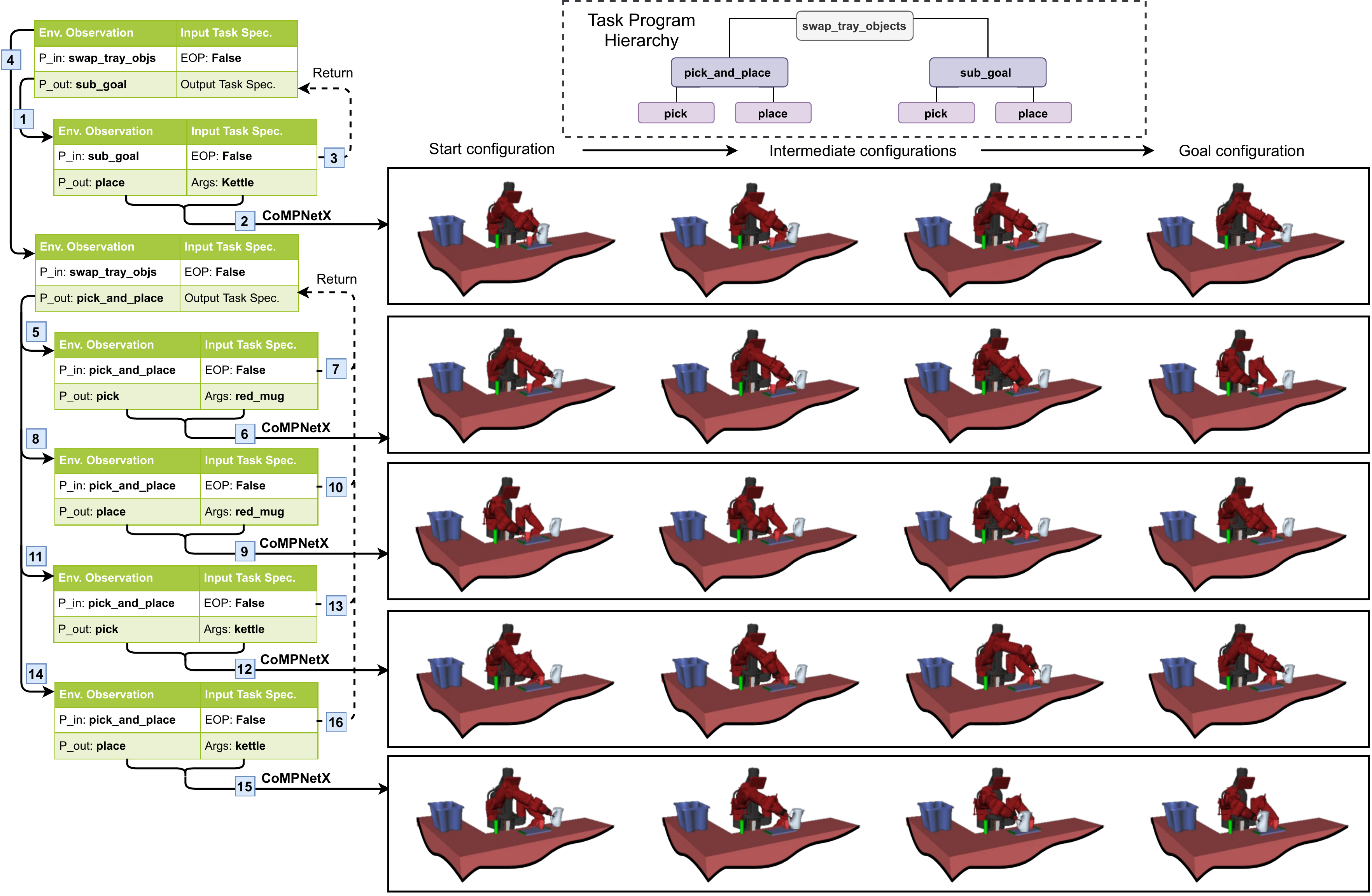}\vspace*{-0.0in}
    \caption{CoMPNetX generating unconstrained and constrained motion sequences to swap red mug and kettle by leveraging the neural task representations given by a learning-based task programmer in the Bartender scenario 3. The task programmer follows the indicated program hierarchy, which also requires subgoal generation for one of the objects as other objects occupy their target configurations. For brevity, we exclude the steps providing next environment observations to the task programmer.  
}\label{bar_sc3}
\end{figure*}

Fig. \ref{bar_sc2} and Fig.\ref{bar_sc3} show the joint execution of NTP2 and CoMPNetX in one of the cases in bartender scenarios 2 and 3, respectively. In this particular case of scenario 2, the task is to move the soda can out of the red mug's target location and then move the red mug to its desired place. Likewise, in this scenario 3 example, the robot has to swap both red mug and kettle locations, which represents a situation where both start and goal locations of the objects are occupied. It can be seen that cross-fertilization of neural task programmers and neural motion planners are crucial for solving challenging practical problems, and CoMPNetX with neural task representation exhibits generalization to problems outside the domain of its training set.


\section{Discussion}
In this section, we briefly discuss the CoMPNetX stochastic behavior and its benefits in learning-based motion planning and the way our approach retains the completeness and optimality guarantees of an underlying SMP planner.
\begin{table}[t]
\centering 
{
\begin{tabular}{ccc}\toprule
\multirow{2}{*}{Datasets}&\multicolumn{2}{c}{Planning times and success rates}\\\cmidrule{2-3}
&\multicolumn{1}{c}{Pick (Unconstrained)}&\multicolumn{1}{c}{Place (Constrained)}\\\midrule


\multirow{1}{*}{Bartender-SC2}& \multirow{1}{*}{$8.57 \pm 5.18$ $ (88.7\%)$} & \multirow{1}{*}{$17.47 \pm 10.07$ $(93.1\%)$} \\ 

\multirow{1}{*}{Bartender-SC3}& \multirow{1}{*}{$6.42 \pm 3.21$ $(79.3\%)$} & \multirow{1}{*}{$10.23 \pm 3.17$ $(81.2\%)$}\\ \bottomrule
\end{tabular}}
\caption{The total mean computation times with standard deviations and mean success rates are presented for CoMPNetX, solving both unconstrained and constrained motion planning problems with underlying MPNet and RRTConnect algorithms, respectively, in the Bartender scenarios 2 and 3.} \label{tab4}\end{table}

CoMPNetX, like its predecessors, uses Dropout \cite{srivastava2014dropout}, with fifty percent probability, to skip some neurons in the generator network $G_{\boldsymbol{\phi}}$ during a forward pass for planning. This process induces a stochastic behavior as in each forward pass, the underlying SMP planner gets a randomly sliced version of CoMPNetX's neural generator. In \cite{gal2016dropout}, it is shown that Dropout-based slicing of neural networks can model uncertainty in the estimation of their parameters. And our studies show that Dropout can help generate a variety of samples in the subspace of a given configuration space that potentially contains a path solution (Fig. \ref{sphere2}). Thus, our framework exploits CoMPNetX stochastic behavior to generate informed trees for any underlying SMP planner in the region that potentially contains a path solution for a given problem through our bidirectional and batch planning methods.

An SMP algorithm $AL$ is \textit{probabilistically complete} if the probability of finding a path solution, if one exists, approaches to $1$ as the number of samples $n$ in their graph $\mathcal{T}_n$ approaches to $\infty$. The primary reason for SMP methods to exhibit such completeness is based on their configuration-space sampling strategies that explore the entire space as the number samples $n$ in their graph grows to a large value. There exist a wide range of SMP algorithms that exhibit \textit{probabilistic completeness} and can be merged with projection and continuation-based constraint-adherence approaches for CMP. In proposition 1, we propose that any SMP method that has \textit{probabilistic completeness} guarantees with their traditional sampling techniques will still retain them with our CoMPNetX sampling strategies, i.e.,  
\\\\ 
\textbf{Conjecture 1 (Probabilistic Completeness)}\textit{ Given a planning problem $\{\boldsymbol{q}_{init},\mathcal{Q}_{goal}, \mathcal{X}_{obs}, \mathbf{F}\}$, and a collision-checker, CoMPNetX will generate samples for an underlying SMP method AL, leading to a tree $\mathcal{T}^{AL}_n$ originating at $\boldsymbol{q}_{init}$ with number of nodes $n$, such that the probability of  finding a path solution $\sigma:[0,1] \mapsto \mathcal{M}_{free}$, if one exists, approaches to one as $n \rightarrow \infty$, i.e., $\mathbb{P}_{n \rightarrow \infty}(\mathcal{T}^{AL} \cap \mathcal{Q}_{goal} \neq \varnothing)=1$.}\\ \\
In addition to \textit{probabilistic completeness}, some SMP algorithms also demonstrate \textit{asymptotic optimality}, i.e., as the number of nodes $n$ in the tree $\mathcal{T}_n$ approaches a large value/infinity, the algorithm will find an optimal path with respect to a given cost function $J(\cdot)$, if one exists, with a probability of 1. In the following proposition 2, we claim that with CoMPNetX, the underlying SMP planner will continue to have their \textit{asymptotic optimality} guarantees, i.e.,\\\\ 
\textbf{Conjecture 2 (Asymptotic Optimality)}\textit{ Given a planning problem $\{\boldsymbol{q}_{init},\mathcal{Q}_{goal}, \mathcal{X}_{obs}, \mathbf{F}, J\}$, and a collision-checker, CoMPNetX adaptively generates configuration samples for an asymptotic optimal SMP algorithm such that the solution, if one exists, asymptotically converges to an optimal path solution, $\sigma^*:[0,1] \mapsto \mathcal{M}_{free}$, w.r.t $J(\cdot)$, as the number of generated samples $n$ approaches to infinity.}\\ \\
\textbf{Justifications for Conjectures 1 \& 2} In algorithms \ref{algo:btree} and \ref{algo:bitree}, our procedures to merge CoMPNetX into any SMP planner consist of exploitation and exploration stages. In the former stage, the procedure uses CoMPNetX to adaptively sample the subspace of an implicit manifold configuration space that potentially contains a path solution. And in the latter, it leverages classical sampling techniques that guarantee uniform coverage of the underlying manifolds. These two stages are balanced through a hyperparameter $N_{ismp}$, which defines the number iterations for which our process relies on exploitation before switching to the exploration stage. Therefore, CoMPNetX with underlying SMP methods do explore the entire configuration spaces as $n \rightarrow \infty$, i.e., $n >> N_{ismp}$. Since CoMPNetX does not alter the internal mechanism of an underlying SMP algorithm and explore the entire state-space over time, it inherits the characteristics of that SMP method, including their \textit{probabilistic completeness} and \textit{asymptotic optimality} and the proofs for our claims remain almost the same as derived in \cite{kingston2019exploring}.

\section{Conclusions \& Future Works}
This paper introduces Constrained Motion Planning Networks X (CoMPNetX), a conditional neural generator-discriminator based path sampling algorithm with neural-gradient based projections to the implicit constraint manifolds. CoMPNetX can speed-up a wide range of SMP planners requiring unidirectional, bidirectional, or batch sampling, thanks to its stochastic behavior and innate capacity of neural networks for parallelization. We also show that the CoMPNetX modular structure naturally allows coupling with learning-based task planners, forming a mutual symbiotic relationship to efficiently solve task and motion planning problems.  Our experimental results validate that CoMPNetX with any underlying SMP approach solves both constrained and unconstrained complex planning problems with high success rates and significantly lower computation times than existing state-of-the-art methods while retaining the worst-case theoretical guarantees.

In our future studies, we plan to incorporate dynamical constraints into CoMPNetX and leverage its symbiotic relationship with learning-based task-programmers and perception methods to address real-world assistive robotic tasks and fully harness its fast, almost real-time computational speed.
\section*{Acknowledgments}
We thank Dmitry Berenson and Frank Park for their insightful discussions and sharing their algorithms' implementations.

\bibliographystyle{IEEEtran}
\bibliography{reference}
\nocite{*}
\begin{appendix}
This section describes our models' architectures with their training details, and other related hyperparameters. We implement and train neural-network models of CoMPNetX and NTP2 using PyTorch \cite{paszke2019pytorch} and port them to C++ via TorchScript to incorporate them with a wide range of OMPL \cite{kingston2019exploring} \cite{sucan2012the-open-motion-planning-library} CMP algorithms. All simulations were designed in OpenRave, and we used standard C++ based OMPL implementations of the benchmark methods.

\subsection{NTP2 Architecture}
\begin{itemize}
    \item Workspace Observation Encoders: The current and target observations are embedded using their encoders, each of which is a multi-layer perceptron. For the Bartender and Kitchen environment, the current and target observation encoders comprise an input layer followed by two hidden layers and an output layer. The 1st hidden layer contains 64 neurons. The 2nd hidden layer contains 128 neurons. The output layer contains 128 neurons. The PReLU non-linearity \cite{trottier2017parametric} follows the hidden layers. A Dropout layer with probability=0.2 is applied after the 1st hidden layer. 

\item Program Embedding Matrix: A program $p$ is represented in the form of a one-hot encoding. A dense representation of a program is learned using an embedding weight matrix of size $7 \times 7$. 

\item Program Planner: The program planner takes the integrated input $\boldsymbol{Z}_p$ comprising the current program embedding with current and target observation encodings to predict the next program $\boldsymbol{p}_{t+1}$ and the end of task probability $r$. It is a multi-layer perceptron with two hidden layers, each with 128 neurons.  The PReLU non-linearity follows the hidden layers. The output of the second hidden layer is the integrated feature vector $\boldsymbol{h}_p$. The integrated feature vector $\boldsymbol{h}_p$ is fed as input to a Program Network and a Terminate Network to obtain the next program and the end of task probability for termination, respectively. The Program Net contains a linear layer with 7 neurons. The Terminate Net has a linear layer with 2 neurons. The outputs of the Program Net and the Terminate Net are encoded using the softmax activation function. 

\item Graph Encoder: The graph encoder takes the program hierarchy and gets embedding matrix $\boldsymbol{H}_p=\{\boldsymbol{h}_i\}^{N_p}_{i=0}$ for each program $\boldsymbol{p}_i$, where $i=(1,2,\cdots,N_p)$ with $N_p$ corresponding to the number of levels in the hierarchy. The features $\boldsymbol{H}_p$ are passed through a linear model to obtain the weights $\boldsymbol{\alpha}_p$. The vector $\boldsymbol{\alpha}_p=(\alpha_1, \alpha_2, ..., \alpha_{N_p})$ is encoded by a softmax-activation layer. Finally, an inner product of the vectors $\boldsymbol{H}_p= (\boldsymbol{h}_1, \boldsymbol{h}_2, ..., \boldsymbol{h}_{N_p})$ and $\boldsymbol{\alpha}_p=(\alpha_1, \alpha_2, ..., \alpha_{N_p})$ is taken to form an output feature vector $\boldsymbol{Z}_g$ representing a weighted program graph.
\item API Decoder: It takes a vector $\boldsymbol{Z}_d$, comprising the current and target observation encodings, graph embedding $\boldsymbol{Z}_g$, and an API-program embedding, as an input. This input vector of size $270$ is passed through a multi-layer perceptron with two hidden layers of size $256$ and $128$ followed by PReLU to obtain the argument $\boldsymbol{a}$.
\end{itemize}

\subsection{CoMPNetX Architecture}
\begin{itemize}
    \item Task Encoder: We use task encoder only for the bartender and kitchen environments. It is a multi-layer perceptron that takes the fixed-size neural task representations $\boldsymbol{Z}_s=[\boldsymbol{Z}_d,\boldsymbol{a}]$ from NTP2 as an input and transforms them to an encoding $\boldsymbol{Z}_c=128$. The structure contains two hidden layers, each with $128$ units, and a non-linear PReLU layer. 
    \item Scene Encoder: The input to the scene encoder is a voxel map of size $40 \times 40 \times 40$, $33\times33\times33$, and $32\times32\times32$ in the sphere, bartender, and kitchen environments, respectively. These maps are transformed into voxel patches for processing with a 2D Convolutional Neural Network (CNN). We use the same network design for all environments except that we have a fever number of neurons in fully-connect layer in the sphere environment. Hence, the model architecture in the sphere planning is reported in the brackets alongside the other environments' model details. The first layer is a 2D-CNN followed by a PReLU, which takes 40, 33, or 32 channels and transforms them to 16 (16) feature maps using a $5\times5$ kernel with $2\times2$ stride. The second layer is again a 2D-CNN followed by PReLU and Max2D pooling with $2\times2$ kernel. In this, the 2D-CNN layer transforms $16 (16)$ feature maps to $8 (8)$ using $3\times3$ kernel with $1\times1$ stride. The output from the second layer is stacked and passed through a sequential neural-network comprising two fully-connected layers. The first layer takes the stacked features and transforms them to $256 (128)$ hidden units followed by a PReLU. The second is an output layer that takes the $256 (128)$ units and transforms them to the output size $\boldsymbol{Z}_o=256 (128)$.    
\item Neural Generator: We use the same generator network architecture for all environments. It is a 6-layer (including an input and output layer) deep neural network. The input is given by concatenating the task encoding $\boldsymbol{Z}_c$ (not included in the sphere environment), scene encoding $\boldsymbol{Z}_o$, and the start $\boldsymbol{q}_0$ and goal $\boldsymbol{q}_T$ configurations. For instance, in the bartender/kitchen environment, the input vector has size of $128+256+13+13=410$, where $13$ corresponds to $7$ DOF of the robot arm and $6$ TSR \cite{berenson2011task} virtual link values. Each of the first four layers is a sandwich of a linear layer, PReLU, and Dropout(0.5) \cite{srivastava2014dropout}. The layers one to four transform the input vectors to 896, 512, 256, and 128 hidden units. The fifth layer does not use the Dropout and transforms the inputs into 64 hidden units. The sixth layer is an output layer and transforms the given vector to given implicit manifold configuration dimensions.
\item Neural Discriminator: We use the same discriminator network architecture for all environments. It is a 3-layer (including an input and output layer) deep neural network. The input is given by concatenating the task encoding $\boldsymbol{Z}_c$, scene encoding $\boldsymbol{Z}_o$, and configuration $\boldsymbol{q}$ and the output is a distance of given configuration from the constraint manifold. The input vector is transformed to the distance $\mathcal{R}^1$ as: $\mathrm{input} \rightarrow 256 \rightarrow 256 \rightarrow 1$ with its first two layers also having a PReLU.
\end{itemize}

\subsection{Training Details}
\begin{itemize}
\item Neural Generator: We train the task and scene encoders together with the neural generator in an end-to-end manner using the mean-square loss against the demonstration trajectories with the Adagrad \cite{duchi2011adaptive} optimizer and the learning rate was set to be $0.01$.
\item Neural Discriminator: We train this module only for the manipulation tasks using mean-square error with Adagrad optimizer and $0.01$ learning rate. 
\item NTP2: The program planner and the API decoder were trained separately with their encoders, each using the cross-entropy loss and the Adagrad optimizer with a learning rate of 0.0001. 
\end{itemize}

\subsection{Constraint-adherence hyperparameters}
\begin{itemize}
\item Neural Projection Operator ($\mathrm{NProj}$): In this operator, the parameters are $(\gamma=0.1, \nu=0.1)$ and $(\gamma=1.2, \nu=0.01)$ for juice-can/soda-can/fuze-bottle and RedMug/BlackMug/Pitcher/Kettle/Door, respectively in the bartender and kitchen environments.
\item Projection: In this operator, we set the tolerance $\varepsilon$ and maximum projection $N$ iterations to $0.001$ and $50$, respectively. The $\lambda_1$ and $\lambda_2$ are set to $2.0$ with step size $\gamma=0.05$. 
\item Continuation: The tolerance and maximum projection iteration values are same as in the projection operator. The rho is $1.5$ and $1.0$ for the bartender and kitchen environments, respectively. The values of epsilon, alpha, exploration (probability of sampling outside the chart), and maximum number of charts are $0.01$, $\pi/6$, $0.9$ (only for classical SMP methods), $5000$, respectively. The value of $\lambda_1=\lambda_2=2.0$ with step size $\gamma=0.05$.
\end{itemize}
\end{appendix}
\end{document}